\documentclass[10pt,twocolumn,letterpaper]{article}

\usepackage[pagenumbers]{cvpr} %

\usepackage{float}
\usepackage[fixed]{fontawesome5}

\usepackage{tabularx}
\usepackage{booktabs}
\usepackage{graphicx}
\usepackage{makecell}
\usepackage{adjustbox}
\usepackage{amssymb}

\def\trex{\scalebox{0.05}{\includegraphics{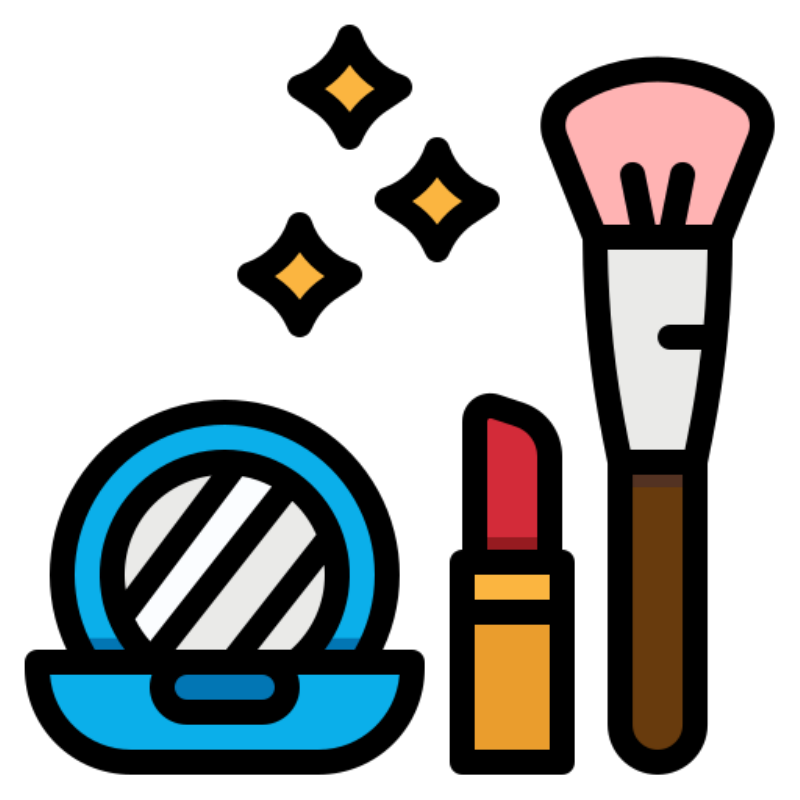}}}  

\newcommand{\name}{\textsc{GenMAC}\xspace}
\newcommand{\design}{\textsc{Design}}
\newcommand{\generation}{\textsc{Generation}}
\newcommand{\redesign}{\textsc{Redesign}}

\newcommand{\designagent}{design agent}
\newcommand{\alignmentagent}{verification agent}
\newcommand{\suggagent}{suggestion agent}
\newcommand{\refineagent}{correction agent}
\newcommand{\structuredagent}{output structuring agent}

\definecolor{cvprblue}{rgb}{0.21,0.49,0.74}
\usepackage[pagebackref,breaklinks,colorlinks,allcolors=cvprblue]{hyperref}

\usepackage{multirow}
\usepackage{epsfig}
\usepackage{colortbl}
\usepackage{xcolor}
\definecolor{mycolor_blue}{HTML}{E7EFFA}
\definecolor{mycolor_green}{HTML}{E6F8E0}
\definecolor{mycolor_gray}{HTML}{ECECEC}
\definecolor{mycolor_red}{HTML}{FFE6E6}
\definecolor{mycolor_yellow}{HTML}{FFFFCC}
\definecolor{mycolor_purple}{HTML}{EBE2F1}
\definecolor{color1}{HTML}{FDFFFD}
\definecolor{color2}{HTML}{FCFEFF}
\definecolor{color3}{HTML}{FFF9E4}
\definecolor{color4}{HTML}{FFF0F5}

\usepackage{placeins}

\def\paperID{1995} %
\def\confName{CVPR}
\def\confYear{2025}

\title{\trex~\name: Compositional Text-to-Video Generation with Multi-Agent Collaboration}

\author{Kaiyi Huang$^{1}$
\qquad
Yukun Huang$^{1}$
\qquad
Xuefei Ning$^{2}$
\qquad
Zinan Lin$^{3}$
\\
Yu Wang$^{2}$
\qquad
Xihui Liu$^{1\dag}$
\vspace{0.2cm}
\\
{\normalsize $^{1}$ The University of Hong Kong \quad $^{2}$ Tsinghua University}
{\normalsize \quad $^{3}$ Microsoft Research}\\
{\normalsize \qquad Project Page: {\href{https://karine-h.github.io/GenMAC/}{https://karine-h.github.io/GenMAC/}}}
}

\begin{document}

\twocolumn[{
\renewcommand\twocolumn[1][]{#1}
\maketitle

\begin{center} 
\vspace{-2em}
    \includegraphics[width=\linewidth]{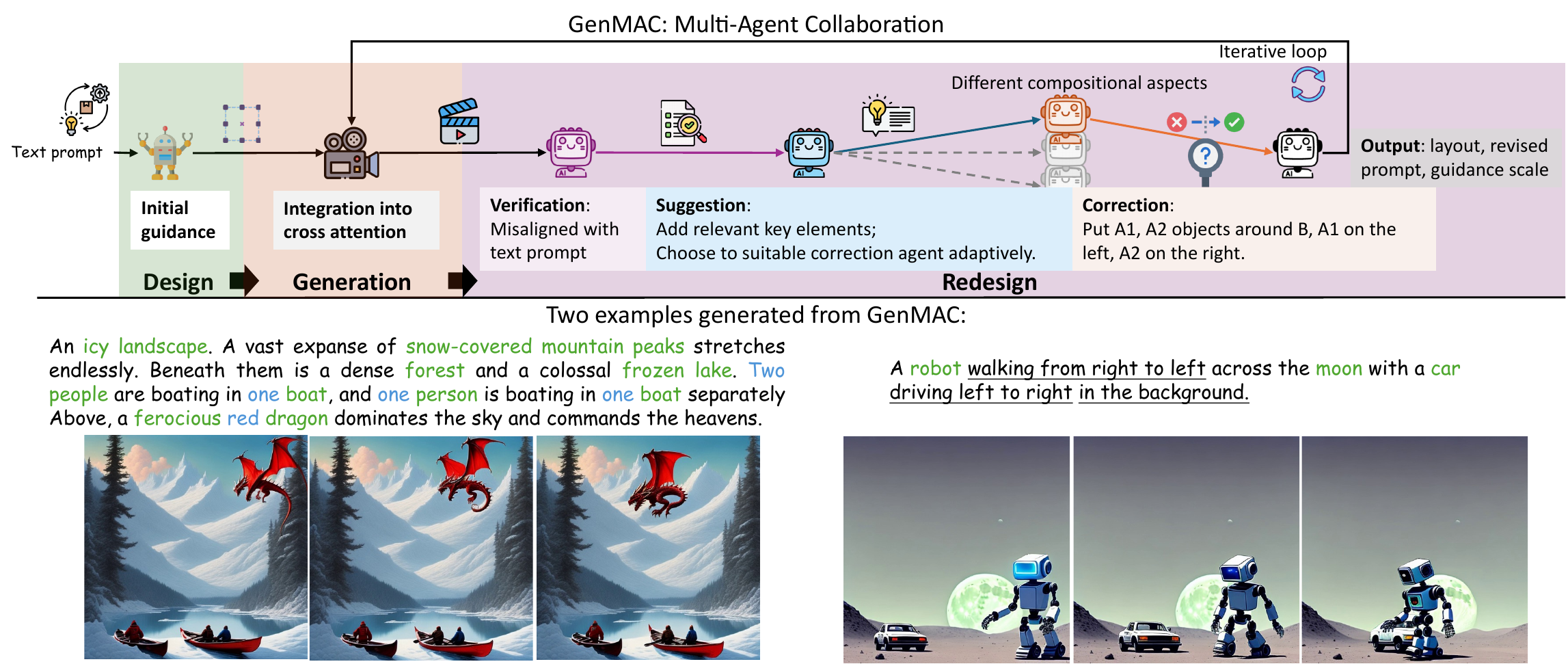}
    \captionsetup{type=figure}
    \vspace{-1em}
    \caption{
   The first row illustrates our multi-agent collaboration approach, showcasing the collaborative workflow, task decomposition in the \redesign{} stage, and adaptive self-routing for \refineagent s.
   The second row presents videos generated by \name{} based on \textbf{complex compositional prompts} involving \textbf{multiple objects, attribute binding, quantity, and dynamic motion binding}. 
   }
    
    \label{fig:teaser}
\end{center}
}]

\let\thefootnote\relax\footnotetext{$^{\dag}$ Corresponding author.}

\begin{abstract}

Text-to-video generation models have shown significant progress in the recent years. However, they still struggle with generating complex dynamic scenes based on compositional text prompts, such as attribute binding for multiple objects, temporal dynamics associated with different objects, and interactions between objects.
Our key motivation is that complex tasks can be decomposed into simpler ones, each handled by a role-specialized MLLM agent. Multiple agents can collaborate together to achieve collective intelligence for complex goals. In this paper, We propose \name{}, an iterative, multi-agent framework that enables compositional text-to-video generation. The collaborative workflow includes three stages: \design{}, \generation{}, and \redesign{}, with an iterative loop between the \generation{} and \redesign{} stages to progressively verify and refine the generated videos. The \redesign{} stage is the most challenging stage that aims to verify the generated videos, suggest corrections, and redesign the text prompts, frame-wise layouts, and guidance scales for the next iteration of generation. To avoid hallucination of a single MLLM agent, we decompose this stage to four sequentially-executed MLLM-based agents: verification agent, suggestion agent, correction agent, and output structuring agent. Furthermore, to tackle diverse scenarios of compositional text-to-video generation, we design a self-routing mechanism to adaptively select the proper correction agent from a collection of correction agents each specialized for one scenario.
Extensive experiments demonstrate the effectiveness of \name{}, achieving state-of-the art performance in compositional text-to-video generation. 
\end{abstract}
    
\section{Introduction}
\label{sec:intro}

With the rapid development of diffusion models~\cite{ho2020denoisingdiffusionprobabilisticmodels, song2021scorebasedgenerativemodelingstochastic, sohldickstein2015deepunsupervisedlearningusing}, text-to-video~\cite{blattmann2023align, blattmann2023stable, ho2022imagen, singer2022make, wu2021godiva,wu2022nuwa,hong2022cogvideo,villegas2022phenaki, zhou2022magicvideo, khachatryan2023text2video, luo2023videofusion, he2022latent, wang2023modelscope} generation has achieved impressive advancements in creating compelling visual content. 
However, current models face significant challenges when tasked with compositional text-to-video generation, particularly in scenarios involving complex spatiotemporal dynamics. 
Accurately generating videos following text prompts that capture intricate compositions, such as multiple objects, attribute binding, diverse actions, and interactions over time, remains a challenging problem~\cite{huang2024t2i, sun2024t2vcompbench}. %

Unfortunately, existing techniques~\cite{videoteris,yang2024vico, wu2023selfcorrectingllmcontrolleddiffusionmodels, wang2024genartistmultimodalllmagent} fall short in following complex text prompts for compositional text-to-video generation. Previous \emph{single-pass} approaches~\cite{videoteris, yang2024vico} generate an entire video in a single pass based on a text prompt. Due to the complexity of compositional prompts, the videos generated with a single pass often miss critical contextual details and therefore fail to follow the text prompts. On the other hand, there have been efforts to introduce self-correction mechanisms to text-to-image generation~\cite{wang2024genartistmultimodalllmagent, wu2023selfcorrectingllmcontrolleddiffusionmodels}, where a Multimodal Large Language Model (MLLM) is used as a \emph{single-agent} to correct misalignments. However, extending such single-agent approaches to video generation presents unique challenges: %
(1) %
Considering the increased complexity from images to videos, current MLLMs are not capable of such complex tasks involving accurate visual understanding, multi-step reasoning, and task planning, and may lead to severe hallucinations. %
(2) 
Different prompts and generated videos require diverse capabilities to handle distinct compositional aspects, for example, consistency across the video, temporal dynamics, spatial dynamics, \etc. Relying on a single predefined agent limits the flexibility and generalizability of self-correction. %

\emph{Our key insight is that even though individual agents may not be capable of complex tasks, task decomposition and role specialization can enable multi-agent collaboration to achieve collective intelligence for complex goals}. Inspired by this, we propose two principles to build the multi-agent collaboration system for compositional text-to-video generation: (1) Complex tasks that require multiple steps of observing, understanding, reasoning, and planning, can be decomposed into sequentially-executed simple tasks.  (2) Considering the complexity in compositional video generation, where different prompts, video outputs, and refinements may be needed, different scenarios may require different role-specialized ``expert'' agents to handle. The proper expert agents should be selected adaptively based on the current scenarios and requirements.

Inspired by these insights and principles, we propose an \emph{iterative}, \emph{multi-agent} framework for compositional text-to-video generation. Our overall collaborative workflow is an iterative process is composed of three stages: \design, \generation, and \redesign, enabling progressive and effective self-corrections over time. 
The \design{} stage leverages an MLLM to establish a high-level structure, determining object layout across frames based on the text prompt.
The \generation{} stage leverages an off-the-shelf video genereation model conditioned on text prompts and layout controls to synthesize videos using the designed layout and and tet prompts.
The \redesign{} stage verifies alignment between the generated video and the text prompt, making necessary adjustments to the design of objects, layouts, or prompts for the next iteration of generation. The \redesign{} stage and \generation{} stage are executed in an iterative loop alternately.

The \redesign{} stage is the most challenging one which requires accurate understanding of videos contents, semantic reasoning of spaitial-temporal dynamics, and planning for the correction and refinement in the next generation iteration. %
Thus, we decompose the \redesign{} stage into multiple sequential tasks - verification, suggestion, correction, and output structuring, executed by different specialized expert agents. %
Furthermore, to handle the complex scenarios of generated videos, text prompts, and refinement needs, %
we design a suite of specialized correction agents for correcting the designs from the perspectives of consistency, temporal dynamics, and spatial dynamics, respectively. A self-routing mechanism is introduced to adaptively select the suitable agent for the current scenario.

To the best of our knowledge, we are the first to address the challenging task of compositional text-to-video generation with multi-agent collaboration. Our core insight is task decomposition and role specialization for multi-agent collaboration and collective intelligence. We propose~\name, an iterative workflow with \design{}, \generation{}, and \redesign{} stages. In the most challenging \redesign{} stage, we propose the novel sequential task decomposition and adaptive self-routing for specialized agent selection. Extensive experiments demonstrate that our proposed~\name~achieves state-of-the-art performance in compositional text-to-video generation in various aspects, significantly outperforming existing methods.

\section{Related Work}
\label{sec:related}
\textbf{Text-to-Video Generation Models.}
Text-to-video generation~\cite{ho2022imagen, singer2022make, zhou2022magicvideo, khachatryan2023text2video, luo2023videofusion, blattmann2023align, he2022latent, wang2023modelscope, guo2023animatediff} has seen advancements with the development of diffusion models~\cite{ho2020denoisingdiffusionprobabilisticmodels}. More recently, language model-based methods~\cite{villegas2022phenaki, chang2023muse, kondratyuk2023videopoet, yu2023magvit, yu2023language, chang2022maskgit} have enabled large-scale training, leading to significant improvements in generating high-quality videos.

\begin{figure*}[ht]   %
  \centering
   \includegraphics[width=\linewidth]{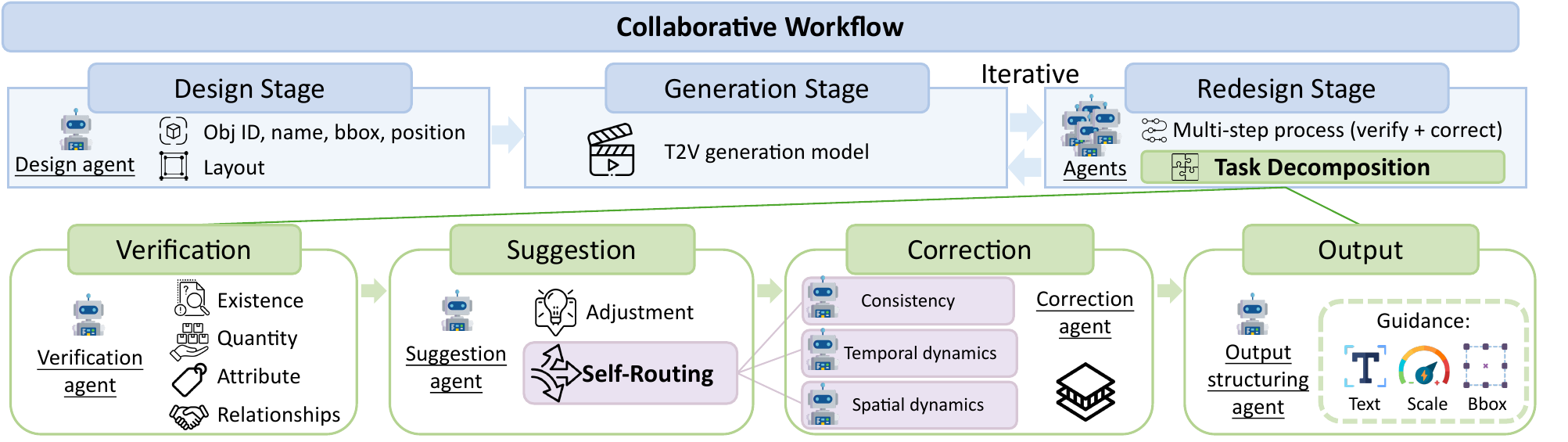}
   \caption{
     \textbf{Framework} of \name{}. \colorbox{mycolor_blue}{Collaborative workflow} includes three stages with an iterative loop: \design, \generation, and \redesign{} (\Cref{sec:workflow}). \colorbox{mycolor_green}{Task decomposition} decomposes the redesign stage into four sub-tasks, handled by four agents: verification agent, suggestion agent, correction agent, and output structuring agent (\Cref{sec:division}). \colorbox{mycolor_purple}{Self-routing} mechanism allows for adaptive selection of suitable correction agent to address the diverse requirements for compositional text-to-video generation (\Cref{sec:self-routing}).
   }
   \label{fig:workflow}
\end{figure*}

\noindent
\textbf{Compositional Text-to-Video Generation.}
There have been studies on compositional text-to-image generation~\cite{liu2022compositional,feng2022training,li2022stylet2i,wu2023harnessing, huang2024t2i, patel2023eclipse, liu2024referee, chefer2023attend,park2021benchmark, lian2023llm, chen2024training,wu2023harnessing, wang2023compositional, meral2023conform, kim2023dense, rassin2024linguistic, gani2023llm, li2023gligen, taghipour2024box, wang2024divide, chen2023reason, yang2024mastering}. 
T2I-CompBench~\cite{huang2024t2i} introduces the first comprehensive benchmark in evaluating compositionality in text-to-image generation models, with attribute binding, relationships, and complex compositions. 
T2V-CompBench~\cite{sun2024t2vcompbench} extends the compositional evaluation to text-to-video generation with the consideration of temporal dimensions. VideoTetris~\cite{videoteris} proposes a framework of spatio-temporal compositional diffusion that enables compositional T2V generation. Vico~\cite{yang2024vico} builds a spatial-temporal attention graph to update the noise latent. 
There exist works that employ an LLM for planning layouts, such as RPG~\cite{yang2024mastering} for text-to-image generation, and LVD~\cite{lian2023llm} and VideoDirectorGPT~\cite{Lin2023VideoDirectorGPT} for video generation. However, the existing works focus on generation in one go, failing to meet complex compositional requirements. 
Our work introduces a collaborative workflow with iterative loop that allows for precise alignment with compositional prompts, progressively refining key elements to achieve greater coherence across spatial and temporal dimensions. 

\noindent
\textbf{LLM-based Agents.}
Recent advancements in (M)LLMs have boosted the development of highly capable AI agents, applied across various domains, such as software development~\cite{wang2024openhandsopenplatformai, qian2024chatdevcommunicativeagentssoftware}, robotics~\cite{driess2023palmeembodiedmultimodallanguage}, scientific research~\cite{tang2024prioritizingsafeguardingautonomyrisks}, society simulation~\cite{park2023generativeagentsinteractivesimulacra}, and beyond. A rapidly growing research focuses on automating interactions with computer environments to solve tasks, such as web manipulation~\cite{yao2023webshopscalablerealworldweb, deng2023mind2webgeneralistagentweb}, gaming~\cite{wang2023voyageropenendedembodiedagent}, command-line coding~\cite{sun2024surveyneuralcodeintelligence}, and text-to-image generation~\cite{wang2024genartistmultimodalllmagent}. 
Various approaches~\cite{park2023generativeagentsinteractivesimulacra, sun2024corexpushingboundariescomplex, wu2023autogenenablingnextgenllm,hong2024metagptmetaprogrammingmultiagent, yuan2024moraenablinggeneralistvideo} have been proposed to enable collaboration and communication among multi-agent to overcome hallucinations. 
While these methods have shown promising results in areas such as automated coding, they often rely on homogeneous agents, limiting the diversity and specialization required for more complex tasks as compositional text-to-video generation. To address these limitations, our work introduces a heterogeneous and hierarchical multi-agent system designed to handle various aspects of compositional requirements in text-to-video generation, expanding the range and effectiveness of multi-agent collaboration in this domain.

\section{Methodology}

Following the principle of task decomposition and role specialization, we introduce \name, a multi-agent framework for compositional text-to-video generation. \name{} follows a three-stage workflow: \design{} $\rightarrow$ \generation{} $\rightarrow$ \redesign{}, with an iterative loop, as outlined in~\Cref{sec:workflow}. Next, we introduce the sequential task decomposition to enable multi-agent collaboration for the most challenging \redesign{} stage in~\Cref{sec:division}. Further, to handle the diverse aspects of design correction, we introduce an adaptive self-routing mechanism to select the most suitable agent for the current situation in~\Cref{sec:self-routing}.

\subsection{Overall Collaborative Workflow}\label{sec:workflow}

\begin{figure*}[ht]   %
  \centering
   \includegraphics[width=\linewidth]{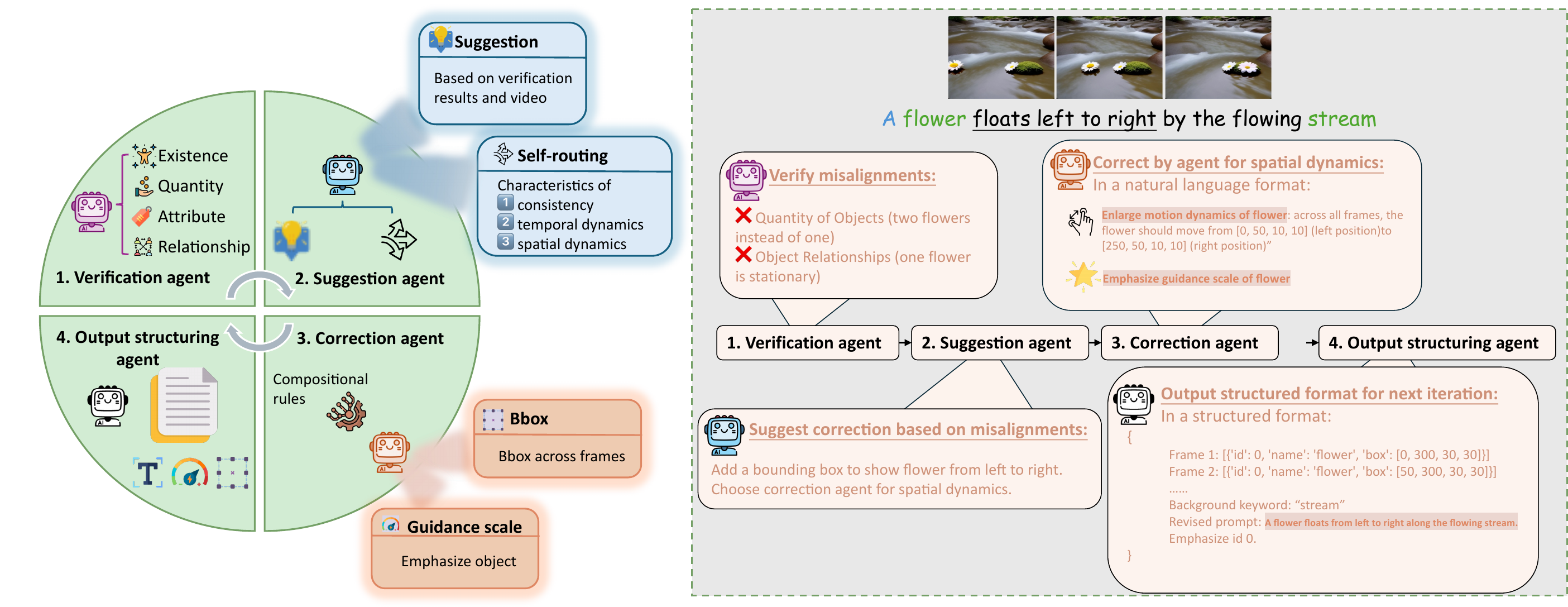}

   \caption{
    \textbf{Illustration} of Task Decomposition for the \redesign{} stage (\Cref{sec:division}). The diagram illustrates the allocation of roles: \alignmentagent, \suggagent, \refineagent, and \structuredagent{} within a sequential task breakdown, highlighting the clear responsibilities of each agent.
   }
   \label{fig:division}
\end{figure*}

Inspired by the human artistic workflow, our multi-agent collaborative framework adopts a \design{} $\rightarrow$ \generation{} $\rightarrow$ \redesign{} pipeline, as shown in~\Cref{fig:workflow}.

\noindent
\textbf{Stage I: \design{}.}
Previous studies have shown that LLMs are able to predict dynamic scene layouts based on text prompts~\cite{lian2023llm}. Inspired by this, our \design{} stage translates the input text prompt into a structured layout, which outlines the key instances, spatial relationships, and temporal dynamics required for compositional video generation. We leverage an LLM to generate structured bounding boxes (which include object IDs, names, box sizes, and positions) for each frame and each instance based on the given text prompt. This stage provides dynamic layout and semantic information to guide the generation stage.

\noindent
\textbf{Stage II: \generation{}.} 
In the \generation{} stage, videos are generated conditioned on the structured layout and guidance scale predicted from the \design{} stage. Specifically, we employ an off-the-shelf text-to-video generation model 
to synthesize visual content that follows the given layout. To incorporate dynamic bounding box control, we follow LVD~\cite{lian2023llm} to inject the structured layout into video diffusion models by guiding the attention maps. %
In this stage, guidance scale is a hyperparameter defined as the scaling coefficient for bounding box guidance in the diffusion process. The guidance scale is initialized as a predefined value in the first generation iteration, and then adjusted by the \redesign{} stages for subsequent generation iterations. %
Details are shown in~\Cref{app: generation_intensity}.

\noindent
\textbf{Stage III: \redesign{}.}
The \redesign{} stage is the core and most challenging stage of our framework. 
It aims to detect misalignment between the generated video and the complex compositional prompt, and adjust the design accordingly for re-generation. We find that a single MLLM agent performs poorly in this complex task. However, if we decompose the final goal into a sequence of simple tasks executed one by one, multiple MLLM agents can collaborate together to produce less hallucinations and more reliable results. We explain the task decomposition and multi-agent collaboration for the \redesign{} stage in~\Cref{sec:division} and~\Cref{sec:self-routing}.

\noindent
\textbf{\redesign{}-\generation{} Loop.}
For complex compositions, a single pass through the workflow may not address all issues in the generated video. Therefore, we introduce an iterative refinement loop between the \generation{} and \redesign{} stages, allowing progressive correction to meet compositional requirements like attribute binding, spatial relationships, and object counts. With guidance from the \redesign{} stage, including bounding boxes, guidance scales, and revised text prompts, the \generation{} stage iteratively improves the video generation results.

\subsection{Task Decomposition for Redesign Stage}\label{sec:division}

The \redesign{} stage requires accurate understanding of the generated videos and text prompts, multi-step reasoning on the video-text misalignments and possible corrections, as well as planning for the new design. We found that this task, especially for videos containing spatial-temporal dynamics, is too difficult for a single MLLM, resulting in hallucinations and inconsistent results (shown in \Cref{tab:ablation}). However, we observe that a single MLLM can be prompted to be an ``expert'' for a specific sub-task, \eg, verifying text-video alignment, or suggesting how to correct the video.
Motivated by the observation, we decompose the \redesign{} stage into sequentially executed easy tasks, each handled by a specialized MLLM-based expert agent (\Cref{fig:division}). %

\noindent
{\small \faTasks} \textbf{Verification Agent} checks how well the video content aligns with the text prompt, focusing on four key aspects: object existence, object quantity, attribute binding, and relationship/interaction. This agent takes the text prompts and generated videos as input, and provides information on the misalignments that need to be addressed in the next generation iteration, \eg, ``There are two flowers in the video, while the text prompt indicates one flower''.

\noindent
{\small \faEdit} \textbf{Suggestion Agent} 
is responsible for suggesting how to refine the design and adaptively selecting the suitable \refineagent{}  (see~\Cref{sec:self-routing}).
The inputs to this agent include the generated videos and the output of the verification agent (\ie, misalignments that need to be corrected).
The outputs of this agent are the suggestions for correction (\eg, ``adding a bounding box for the missing tree'', or ``move the box of apple to the left and make it smaller'') and the selection of suitable correction agent  (see~\Cref{sec:self-routing}).

\noindent
{\small \faTools} \textbf{Correction Agent} is responsible for correcting the bounding box layout, and guidance scale
in the current design. %
The \refineagent{} takes the generated video, current design (bounding box layout and guidance scale), and suggestion from the suggestion agent as inputs. It outputs the corrections to the current design of bounding boxes and guidance scales. For example, ``at frame 1, the apple is at [60, 50, 10, 10]; later, it should move to [30, 50, 10, 10]''. %

\noindent
{\small \faUserCheck} \textbf{Output Structuring Agent} translates the correction results from the \refineagent{} into structured outputs in json format, ready to be passed to the next generation stage. It takes the generated videos and corrections from the correction agent as input, and outputs the new design in json format. The formatted new design includes information of bounding boxes, (revised) text prompts, and guidance scale. %

\subsection{Adaptive Self-Routing for Correction Agents}\label{sec:self-routing}

Considering the complexity of tackling spatial-temporal dynamics and consistency for compositional text-to-video generation, it requires the agent to be capable of handling diverse aspects and making proper decisions based on the current situation. We found that a single predefined agent cannot address all required aspects effectively (\Cref{tab:ablation}). Therefore, we propose a suite of MLLM-based specialized agents, each designed to be an expert for a distinct aspect of video generation. Through an analysis of compositional text-to-video generation, we observe that the most common issues can be categorized into three categories: \textit{consistency}, \textit{temporal dynamics}, and \textit{spatial dynamics}. Therefore, we design the three expert agents for the three perspectives. %

\noindent
\textbf{Correction Agent for Consistency}. For tasks requiring temporal consistency, such as keeping the attribute and spatial layout to be the consistent across the video frames, we introduce a correction agent focusing on maintaining the consistency over time. %

\noindent
\textbf{Correction Agent for Temporal Dynamics}. This agent is designed for cases with temporal dynamics, such as attribute changes or dynamic actions. It adjusts the layout and descriptions dynamically over time, ensuring that each frame reflects the evolving attributes accurately.

\noindent
\textbf{Correction Agent for Spatial Dynamics}. One challenging scenario for compositional text-to-video generation is the change of object locations over time. We design an expert agent for this scenario to handle moving objects in videos. The agent is particularly good at understanding and reasoning dynamic locations and spatial relationships. %

In our self-routing mechanism, the \suggagent{} adaptively selects the appropriate correction agent based on the current generated video and the video-text misalignments that needs to be addressed. %
For example, it routes to the consistency agent to improve the temporal consistency of attributes across the video, and routes to the temporal dynamics agent if the generated video fails to reflect the change of object states over time.
An example is illustrated in the right part of \Cref{fig:division}, where selecting the correction agent for spatial dynamics enables larger motion dynamics, such as the bounding box of the flower moving from the leftmost to the rightmost position. This self-routing process allows \name~to make context-aware, precise corrections by selecting the most suitable agent.

\section{Experiments}
We present the experimental setup in~\Cref{exp:exp setup}, baseline comparisons in~\Cref{exp:quantitative compare} and~\Cref{exp:qualitative compare}, iterative analysis in~\Cref{exp: analysis}, and ablation studies of key components in~\Cref{exp: ablation}.

\begin{table*}[t]

\centering
    \caption{\textbf{Quantitative Comparison on T2V-CompBench.} Compared with existing text-to-video generation models and compositional methods, ~\name~demonstrates exceptional performances in consistent attribute binding, dynamic attribute binding, spatial relationships, motion binding, action binding, object interactions, and generative numeracy, indicating our method achieves superior compositional generation ability. We highlight the best score in \colorbox{mycolor_green}{green}, and the second-best value in \colorbox{mycolor_blue}{blue}.
    The baseline data are sourced from~\cite{sun2024t2vcompbench}. 
    } 
\label{tab:benchmark}
\resizebox{\linewidth}{!}{%
\begin{tabular}%
{lccccccccccccccccccc}
\toprule %
\multicolumn{1}{c}{{\textbf{Model}}} & \multicolumn{1}{c}{\textbf{Consist-attr}} & \multicolumn{1}{c}{\textbf{Dynamic-attr}} & \multicolumn{1}{c}{\textbf{Spatial}} & \multicolumn{1}{c}{\textbf{Motion}} & \multicolumn{1}{c}{\textbf{Action}} & \multicolumn{1}{c}{\textbf{Interaction}} & \multicolumn{1}{c}{\textbf{Numeracy}}    \\
\cmidrule(lr){1-1} \cmidrule(lr){2-2} \cmidrule(lr){3-3} \cmidrule(lr){4-4} \cmidrule(lr){5-5} \cmidrule(lr){6-6} \cmidrule(lr){7-7} \cmidrule(lr){8-8} 
\multicolumn{1}{c}{Metric} & Grid-LLaVA $\uparrow$    & D-LLaVA  $\uparrow$    & G-Dino $\uparrow$ & DOT  $\uparrow$   & Grid-LLaVA  $\uparrow$ & Grid-LLaVA  $\uparrow$ & G-Dino  $\uparrow$  \\
\midrule

ModelScope~\cite{wang2023modelscope} &0.5483 &0.1654 &0.4220 & 0.2552&0.4880 &0.7075 &0.2066 \\
ZeroScope~\cite{Zeroscope} &0.4495 &0.1086 &0.4073 & 0.2319&0.4620 &0.5550 &0.2378 \\
Latte~\cite{ma2024latte} &0.5325 &0.1598 & 0.4476& 0.2187&0.5200 &0.6625 &0.2187 \\
Show-1~\cite{zhang2023show}  &0.6388 &0.1828 & 0.4649& 0.2316&0.4940 &0.7700 &0.1644 \\
VideoCrafter2~\cite{chen2024videocrafter2} &0.6750 &0.1850 & 0.4891& 0.2233&0.5800 &0.7600 &0.2041 \\
Open-Sora 1.1~\cite{opensora} &0.6370 &0.1762 & \cellcolor{mycolor_blue}{0.5671}&0.2317 &0.5480 & 0.7625&0.2363 \\
Open-Sora 1.2~\cite{opensora} &0.6600 &0.1714 & 0.5406& 0.2388&0.5717 &0.7400 &0.2556 \\
Open-Sora-Plan v1.0.0~\cite{pku_yuan_lab_and_tuzhan_ai_etc_2024_10948109} &0.5088 &0.1562 &0.4481 & 0.2147&0.5120 &0.6275 &0.1650 \\
Open-Sora-Plan v1.1.0~\cite{pku_yuan_lab_and_tuzhan_ai_etc_2024_10948109} &\cellcolor{mycolor_blue}{0.7413} &0.1770 &0.5587 &0.2187& \cellcolor{mycolor_blue}{0.6780}&0.7275 &\cellcolor{mycolor_blue}{0.2928} \\
CogVideoX-5B~\cite{yang2024cogvideox} & 0.7220	& 0.2334	& 0.5461	& 0.2943	& 0.5960	& \cellcolor{mycolor_blue}{0.7950}	& 0.2603 \\
AnimateDiff~\cite{guo2023animatediff} &0.4883 & 0.1764 & 0.3883&0.2236 &0.4140 & 0.6550& 0.0884\\
VideoTetris~\cite{videoteris} & 0.7125& 0.2066 & 0.5148& 0.2204&0.5280 & 0.7600& 0.2609\\
Vico~\cite{yang2024vico} & 0.7025 & \cellcolor{mycolor_blue}{0.2376} & 0.4952 & 0.2225 & 0.5480 & 0.7775 & 0.2116 \\
LVD~\cite{lian2023llm} &0.5595 &0.1499 & 0.5469& 0.2699&0.4960 &0.6100 & 0.0991\\
MagicTime~\cite{yuan2024magictime} & - & 0.1834 & - & - & - & - & -\\
Pika~\cite{pika} (Commercial) & 0.6513 & 0.1744 & 0.5043 & 0.2221 & 0.5380 & 0.6625 & 0.2613\\
Gen-3~\cite{gen3} (Commercial) & 0.7045 & 0.2078 & 0.5533 & \cellcolor{mycolor_blue}{0.3111} & 0.6280 & 0.7900 & 0.2169\\
\midrule
\textbf{~\name~(Ours)} & \cellcolor{mycolor_green}{0.7875} & \cellcolor{mycolor_green}{0.2498} & \cellcolor{mycolor_green}{0.7461} & \cellcolor{mycolor_green}{0.3623} & \cellcolor{mycolor_green}{0.7273} & \cellcolor{mycolor_green}{0.8250} & \cellcolor{mycolor_green}{0.5166}\\

\bottomrule
\end{tabular}
}
\end{table*}

\begin{figure*}[ht]   %
  \centering
   \includegraphics[width=0.9\linewidth]{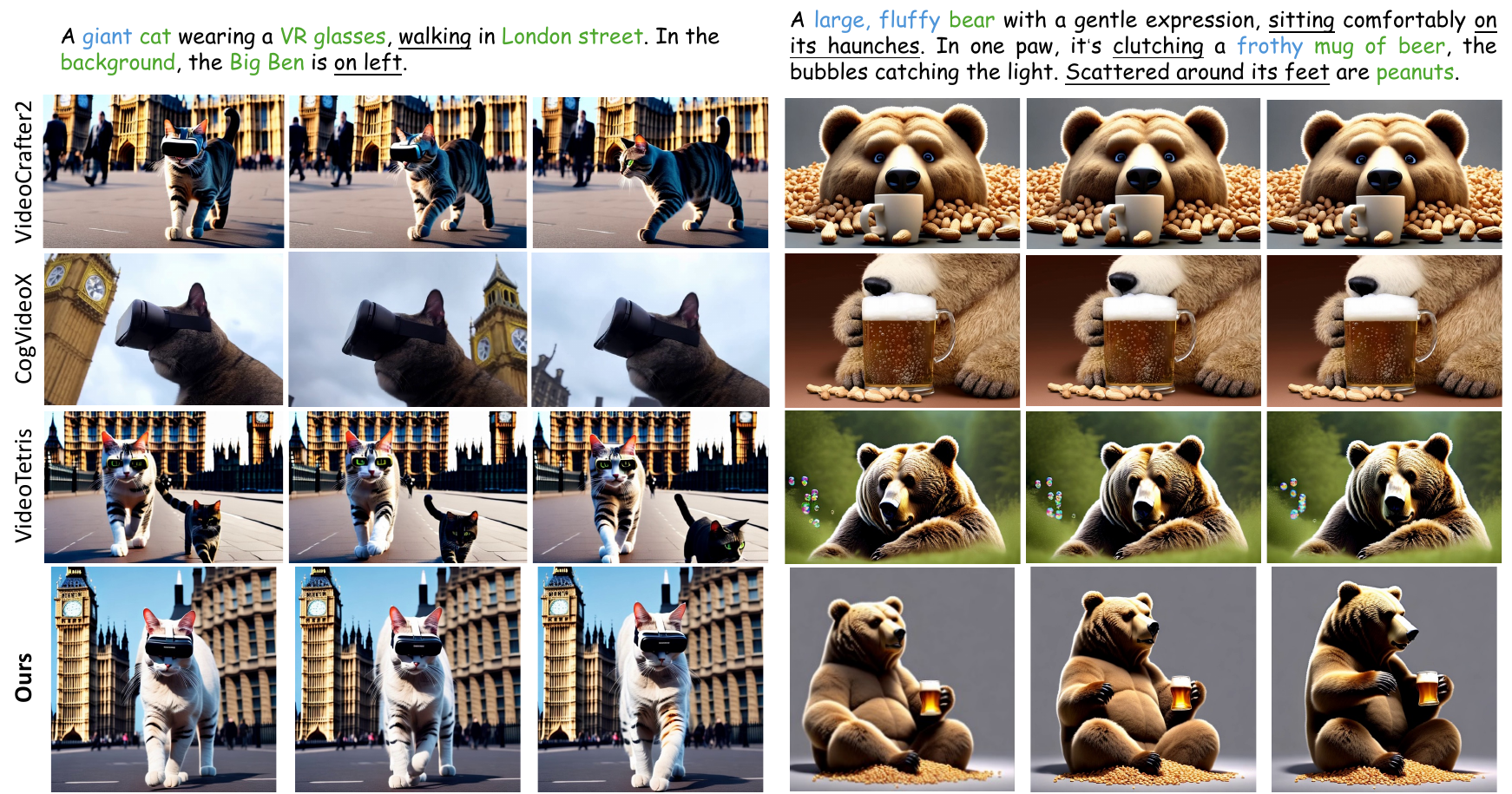}
   \caption{
   \textbf{Qualitative Comparison.} Our proposed~\name~generates videos that accurately adhere to complex compositional scenarios, demonstrating a clear advantage in handling such requirements in comparision with
SOTA text-to-video models.
   }
   \label{fig:qualitative compare}
\end{figure*}

\begin{table*}[t!]
	\centering
	\caption{\textbf{Ablation Study}. The \colorbox{mycolor_green}{complete framework} achieves the highest scores. 
    }\label{tab:ablation}
\centering
\setlength{\tabcolsep}{3.0pt}
\small
\resizebox{\textwidth}{!} {%
\begin{tabular}
{lccccccccccccccccccc}
\bottomrule[1pt]\rowcolor[HTML]{FAFAFA}

\toprule \rowcolor[HTML]{FAFAFA}
\multicolumn{1}{c}{{\textbf{}}} & \multicolumn{1}{c}{\textbf{Consist-attr}} & \multicolumn{1}{c}{\textbf{Dynamic-attr}} & \multicolumn{1}{c}{\textbf{Spatial}} & \multicolumn{1}{c}{\textbf{Motion}} & \multicolumn{1}{c}{\textbf{Action}} & \multicolumn{1}{c}{\textbf{Interaction}} & \multicolumn{1}{c}{\textbf{Numeracy}}    \\
\cmidrule(lr){1-1} \cmidrule(lr){2-2} \cmidrule(lr){3-3} \cmidrule(lr){4-4} \cmidrule(lr){5-5} \cmidrule(lr){6-6} \cmidrule(lr){7-7} \cmidrule(lr){8-8} 
\multicolumn{1}{c}{Metric} & Grid-LLaVA $\uparrow$    & D-LLaVA  $\uparrow$    & G-Dino $\uparrow$ & DOT  $\uparrow$   & Grid-LLaVA  $\uparrow$ & Grid-LLaVA  $\uparrow$ & G-Dino  $\uparrow$  \\
\hline

\toprule[0.8pt]
\rowcolor[HTML]{FDFFFD}  \multicolumn{8}{c}{\textit{Multiple stages and iterative refinement}} \\

\hline

\rowcolor[HTML]{FDFFFD} \generation{}         & 0.6663 & 0.2308 & 0.5106 & 0.2178 & 0.5640 & 0.8125 & 0.2869 \\

\rowcolor[HTML]{FDFFFD} + \redesign{} & 0.7208 & 0.2310 & 0.6680 & 0.2468 & 0.6545 & 0.8000 & 0.2869 \\

\rowcolor[HTML]{FDFFFD}+ iterative & 0.7495	& 0.2402	& 0.7032	& 0.2608	& 0.7060	& 0.8125	& 0.4188\\
\midrule
\rowcolor[HTML]{FDFFFD} \design{} + \generation{} & 0.7045 & 0.2320 & 0.7264 & 0.3327 & 0.6880 & 0.7525 & 0.4113 \\
\rowcolor[HTML]{FDFFFD} + \redesign{} & 0.7513 & 0.2378 & 0.7361 & 0.3474 & 0.7160 & 0.7850 & 0.4794 \\

\hline
\toprule[0.8pt]
\rowcolor[HTML]{FCFEFF}  \multicolumn{8}{c}{\textit{Role specialization in the \redesign{} stage}} \\

\hline

\rowcolor[HTML]{FCFEFF} Single-agent               & 0.7200 & 0.2382 & 0.7336 & 0.3336 & 0.6740 & 0.7700 & 0.3984 \\

\rowcolor[HTML]{FCFEFF} + iterative & 0.7150 & 0.2258 & 0.7336 & 0.3323 & 0.6808 & 0.7700 & 0.3984 \\

\rowcolor[HTML]{FCFEFF} Verification + Correction & 0.7138 & 0.2251	& 0.7134 & 0.3179 & 0.6680	& 0.7125	& 0.4284 \\
\rowcolor[HTML]{FCFEFF} + iterative & 0.7113	& 0.2260	& 0.7149	& 0.3318	& 0.6640	& 0.7686	& 0.4222\\
\rowcolor[HTML]{FCFEFF} Verification + Suggestion + Correction & 0.7370	& 0.2324	& 0.7300	& 0.3173	& 0.7080	& 0.7825	& 0.4469\\
\rowcolor[HTML]{FCFEFF} + iterative & 0.7588	& 0.2440	& 0.7450	& 0.3196	& 0.7184	& 0.8175	& 0.4766 \\

\hline
\toprule[0.8pt]
\rowcolor[HTML]{FFFDF5} 
\multicolumn{8}{c}{\textit{Self-routing for the correction agent}} \\
\hline

\rowcolor[HTML]{FFFDF5} w/o self-routing & 0.7175 & 0.2316 & 0.7391 & 0.3431 & 0.7240 & 0.8025 & 0.4348 \\
\rowcolor[HTML]{FFFDF5} + iterative & 0.7325 & 0.2296 & 0.7408 & 0.3517 & 0.7160 & 0.8150 & 0.4647 \\

\hline

\rowcolor[HTML]{E6F8E0} 
\toprule[0.8pt]

\rowcolor[HTML]{E6F8E0}  \name{} (\textbf{ours}) & \textbf{0.7875} & \textbf{0.2498} & \textbf{0.7461} & \textbf{0.3623} & \textbf{0.7273} & \textbf{0.8250} & \textbf{0.5166} \\

\bottomrule[0.8pt]

\end{tabular}
}
\end{table*}

\subsection{Experimental Setups}
\label{exp:exp setup}
\textbf{Implementation Details.} 
We apply our~\name~on VideoCrafter2~\cite{chen2024videocrafter2} as the backbone for the \generation{} stage to generate videos with 65 frames, 512x512 resolution. 
We use GPT-4o~\cite{gpt-4o} as LLM agent. See more details in~\Cref{app: implementation}.

\noindent \textbf{Evaluated Models.}
We compare our approach with 17 text-to-video generation models, including 15 open-source models and 2 commercial models: ModelScope~\cite{wang2023modelscope}, ZeroScope~\cite{Zeroscope}, Latte~\cite{ma2024latte}, Show-1~\cite{zhang2023show}, VideoCrafter2~\cite{chen2024videocrafter2}, Open-Sora 1.1 and 1.2~\cite{opensora}, Open-Sora-Plan v1.0.0 and v1.1.0~\cite{pku_yuan_lab_and_tuzhan_ai_etc_2024_10948109}, CogVideoX-5B~\cite{yang2024cogvideox}, AnimateDiff~\cite{guo2023animatediff}, VideoTetris~\cite{videoteris}, Vico~\cite{yang2024vico}, MagicTime~\cite{yuan2024magictime}, LVD~\cite{lian2023llm}, Pika~\cite{pika}, and Gen-3~\cite{gen3}.

\noindent \textbf{Benchmark and Evaluation Metrics.} We use T2V-CompBench~\cite{sun2024t2vcompbench} as the benchmark to evaluate the quality of compositional text-to-video generation from seven aspects: consistent and dynamic attribute binding, spatial relationships, motion binding, action binding, object interactions, and generative numeracy.

\subsection{Quantitative Comparisons}
\label{exp:quantitative compare}

We quantitatively compare our~\name~with text-to-video generation models, evaluating seven crucial compositional aspects in~\Cref{tab:benchmark}.
Our~\name~consistently achieves consistently better performance across seven categories than all the 17 baselines. 
Among the baselines, the foundation models such as Open-Sora-Plan~\cite{pku_yuan_lab_and_tuzhan_ai_etc_2024_10948109}, Open-Sora~\cite{opensora}, VideoCrafter2~\cite{chen2024videocrafter2}, CogVideoX~\cite{yang2024cogvideox}, the commercial Gen-3~\cite{gen3}, and the methods specifically designed for compositionality like VideoTetris~\cite{videoteris} and Vico~\cite{yang2024vico}, can achieve higher quality. 
Our method achieves superior performances compositionality, with an exceptional increase in generative numeracy (76.43\% above the second-best), and notable improvements in spatial relationships (31.56\%), motion binding (16.46\%), action binding (7.27\%), consistent attribute binding (6.23\%), dynamic attribute binding (5.13\%), and interactions (4.43\%).

\begin{figure*}[ht]   %
  \centering
   \includegraphics[width=\linewidth]{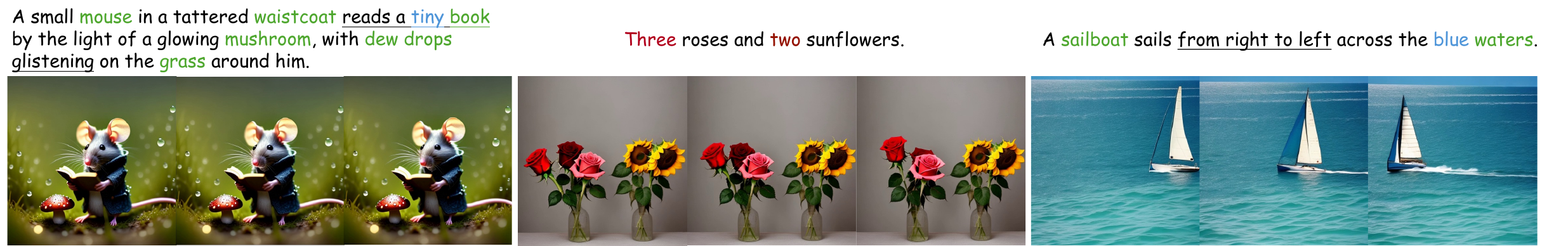}
   \caption{
   \textbf{Qualitative Results.} Our proposed~\name~generates videos that highly aligned with complex compositional prompts, including attribute binding, multiple objects, quantity, and dynamic motion binding.
   }
   \label{fig:qualitative ours}
\end{figure*}

\subsection{Qualitative Comparisons}
\label{exp:qualitative compare}

\noindent
\textbf{Comparison with Existing Methods.}
We show visual comparisons on the video frames of our proposed~\name~and VideoCrafter2~\cite{chen2024videocrafter2}, CogVideoX-5B~\cite{yang2024cogvideox}, and VideoTetris~\cite{videoteris}.
\Cref{fig:qualitative compare} presents the visual comparisons of the video frames generated by \name and existing models, including VideoCrafter2~\cite{chen2024videocrafter2}, CogVideoX-5B~\cite{yang2024cogvideox}, and VideoTetris~\cite{videoteris}.
We can observe that existing models struggle to meet compositional requirements. In the left example, VideoCrafter2~\cite{chen2024videocrafter2} omits the VR glasses, VideoTetris~\cite{videoteris} generates two cats instead of one and misses the VR glasses, and CogVideoX-5B~\cite{yang2024cogvideox} only shows part of the cat near Big Ben. In the right example, VideoTetris~\cite{videoteris} does not depict ``sitting on haunches,'' while VideoCrafter2~\cite{chen2024videocrafter2} and CogVideoX-5B~\cite{yang2024cogvideox} only show partial views of the bear. 
These examples highlight the challenges in compositional text-to-video generation. 
In contrast, our proposed~\name generates videos that accurately adhere to complex compositional scenarios. 
See more examples in~\Cref{app: qualitative_compare}.

\noindent
\textbf{More Qualitative Examples.} 
The results in~\Cref{fig:qualitative ours} show that \name{} demonstrates better performances in compositionality. See more examples in~\Cref{app: qualitative_ours}.

\begin{figure}[ht]   %
  \centering
   \includegraphics[width=\linewidth]{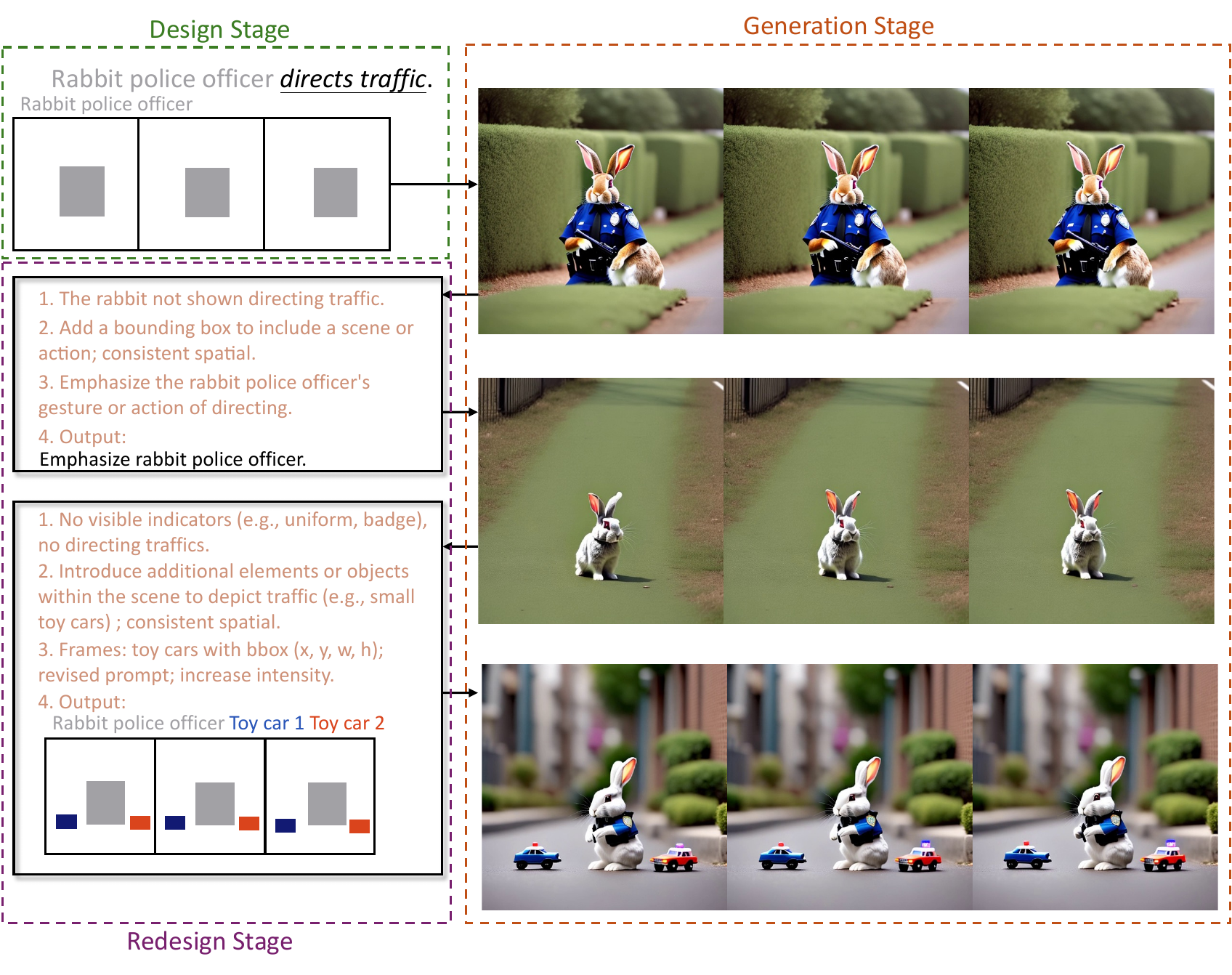}
   \caption{\textbf{Visualization} of the iterative refinement process in our multi-agent framework, demonstrating iterations enhance scene accuracy by progressively aligning video content with compositional prompts.
   }
   \label{fig:qualitative_2}
   \vspace{-10pt}
\end{figure}

\subsection{Analysis on Iterative Generation}
\label{exp: analysis}
\noindent
\textbf{An Example.}
\Cref{fig:qualitative_2} presents an iterative refinement example.
The \designagent{} initially establishes layouts of a rabbit police officer across frames. However, the generated video does not adhere to the ``directing traffic'' element in the prompt. In the first iteration of \redesign{}, the agents identify the misaligned elements and increase the guidance scale for the bounding box of the rabbit police officer. The generated video, however, shows only a rabbit without the necessary traffic context or police uniform. In the second iteration, agents in \redesign{} stage detect these discrepancies and explicitly add elements like toy cars to indicate traffic, while further increasing the the guidance scale for the bounding box of the rabbit police officer. Additionally, this iteration revises the prompt to include the new elements and reinforce the scenario (\textit{i.e.}, ``on the street''). See more examples in~\Cref{app: iteration_results}

\begin{figure}[ht]   %
  \centering
   \includegraphics[width=\linewidth]{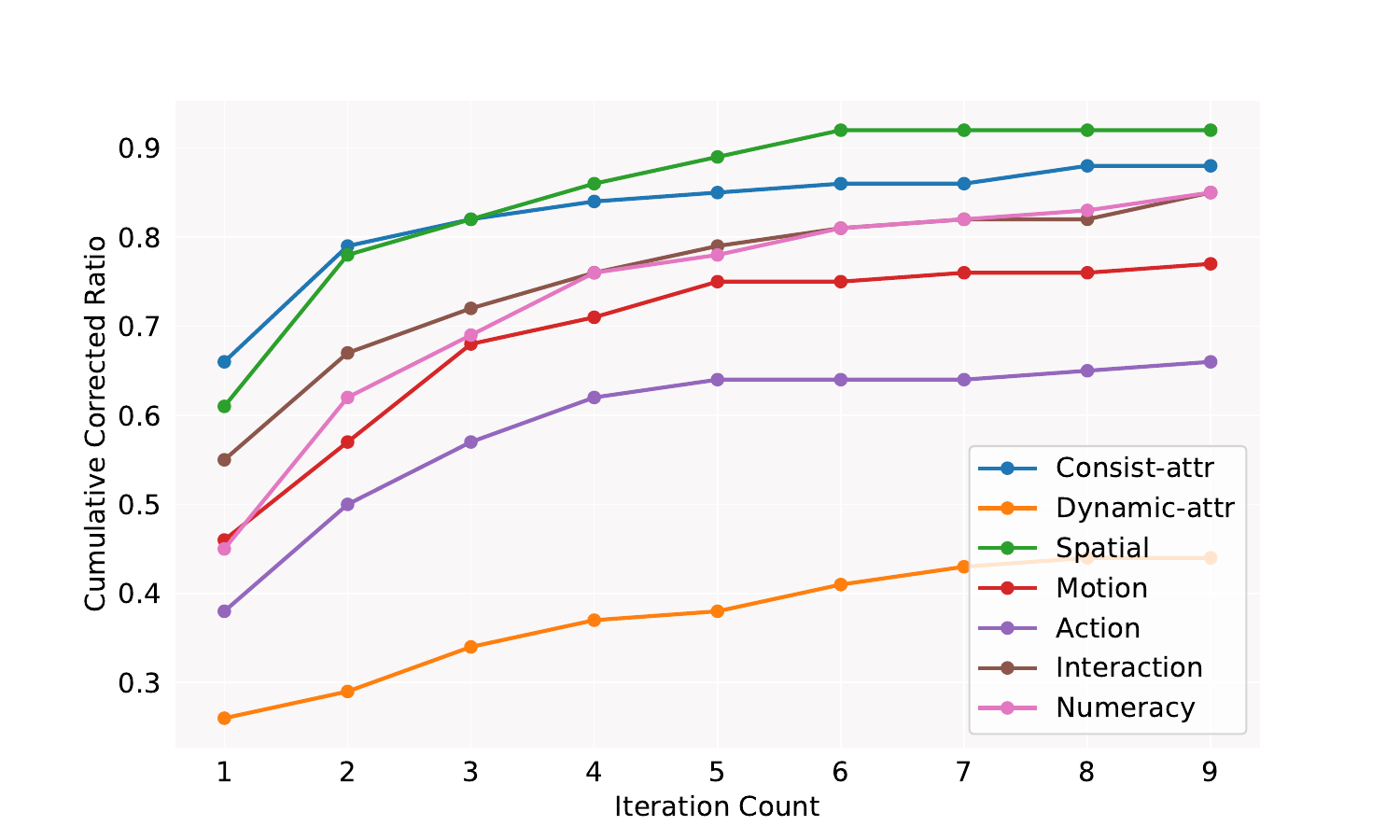}
   \caption{Cumulative Corrected Ratio. For each subset in T2V-CompBench, we calculate the ratio of prompts that have completed the refinement and exited the \name loop to the total size of the subset in each iteration. Dynamic attribute binding remains challenging, while generative numeracy, spatial relationships, and motion binding show substantial improvements from iteration 1 to 9.%
   }

   \label{fig:cdf}
\end{figure}

\noindent
\textbf{Iterative Refinement of Different Compositional Aspects.} 
T2V-CompBench consists of seven subsets of prompts, each emphasizing one of the seven compositional aspects. We calculate the cumulative corrected ratio within each subset at every iteration, which is the ratio of prompts that have completed the refinement and exited the \name loop to the total size of the subset. As shown in~\Cref{fig:cdf}, the corrected ratio gradually increases with iterations across all compositional aspects, demonstrating the necessity of iterative refinement.

Among the seven compositional aspects, we can observe that dynamic attribute binding presents the greatest challenge, consistently showing the lowest corrected ratios across iterations. In contrast, consistent attribute binding and spatial relationships begin with higher corrected ratios.
\name demonstrates particular strengths in enhancing generation quality for certain compositional aspects,  namely numeracy, spatial relationships, and motion binding, with improvements of 40\%, 31\%, and 31\%, respectively,  from iteration 1 to 9.

\subsection{Ablation Study}
\label{exp: ablation}

To evaluate the effectiveness of each design component in \name, we perform three ablation studies in~\Cref{tab:ablation}. 

\noindent
\textbf{Effect of Multiple Stages and Iterative Refinement.} 
The method with only a \generation{} stage yields the lowest generation quality. Introducing a \design{} stage (``\design{}+\generation{}'') improves the quality. Adding one (``+\redesign{}'') or multiple iterative \redesign{} stages (``+iterative'') further enhances the quality.

\noindent
\textbf{Effect of Role Specialization in the \redesign{} Stage.}
Separating roles in the \redesign{} stage significantly enhances the generation quality.
(1) Quantitative results show that the multi-agent design can bring notable improvements over single-agent framework, \textit{e.g.}, the iterative single-agent framework can only achieve 0.715 on consistent attribute binding, much lower than that of \name (0.7875).
(2) Removing the \structuredagent{} and \suggagent{} from the \redesign{} stage leads to significant degradation in quality. For instance, the 2-agent \redesign{} achieves a score of only 0.7113 for consistent attribute binding, compared to 0.7588 for the 3-agent \redesign{}, and 0.7875 for the 4-agent \redesign{} (\name).

\noindent
\textbf{Effect of Self-Routing for the Correction Agent.}
We compare our method with a method version without the self-routing mechanism for the \refineagent{}. In this simplified version, one single \refineagent{} handles information from all compositional aspects. The results clearly highlight the advantage of the self-routing mechanism.

\section{Conclusion}
In this paper, we address the challenges faced by state-of-the-art video generation models in producing complex compositional video content. Specifically, we introduce an iterative, multi-agent framework that enables high-quality compositional generation. Our workflow incorporates iterative refinement and decomposes the task into three manageable stages: \design{}, \generation{}, and \redesign{}. We further decompose the core \redesign{} stage into four sequential tasks executed by specialized agents: verification, suggestion, correction, and output structuring. Finally, we design a self-routing mechanism that adaptively selects among multiple correction agents, enabling better handling of diverse compositional aspects.
Extensive experiment results confirm the effectiveness and superiority of our method in generating compositional text-to-video generation.

{
    \small
    \bibliographystyle{ieeenat_fullname}
    \bibliography{ref}

\begin{thebibliography}{73}
\providecommand{\natexlab}[1]{#1}
\providecommand{\url}[1]{\texttt{#1}}
\expandafter\ifx\csname urlstyle\endcsname\relax
  \providecommand{\doi}[1]{doi: #1}\else
  \providecommand{\doi}{doi: \begingroup \urlstyle{rm}\Url}\fi

\bibitem[Zer(2023)]{Zeroscope}
{ZeroScope}.
\newblock \url{https://huggingface.co/cerspense/zeroscope_v2_576w/}, 2023.
\newblock Accessed: 2024-11-14.

\bibitem[pik(2023)]{pika}
{Pika Art}.
\newblock \url{https://pika.art/}, 2023.
\newblock Accessed: 2024-11-14.

\bibitem[Blattmann et~al.(2023{\natexlab{a}})Blattmann, Dockhorn, Kulal, Mendelevitch, Kilian, Lorenz, Levi, English, Voleti, Letts, et~al.]{blattmann2023stable}
Andreas Blattmann, Tim Dockhorn, Sumith Kulal, Daniel Mendelevitch, Maciej Kilian, Dominik Lorenz, Yam Levi, Zion English, Vikram Voleti, Adam Letts, et~al.
\newblock Stable video diffusion: Scaling latent video diffusion models to large datasets.
\newblock \emph{arXiv preprint arXiv:2311.15127}, 2023{\natexlab{a}}.

\bibitem[Blattmann et~al.(2023{\natexlab{b}})Blattmann, Rombach, Ling, Dockhorn, Kim, Fidler, and Kreis]{blattmann2023align}
Andreas Blattmann, Robin Rombach, Huan Ling, Tim Dockhorn, Seung~Wook Kim, Sanja Fidler, and Karsten Kreis.
\newblock Align your latents: High-resolution video synthesis with latent diffusion models.
\newblock In \emph{Proceedings of the IEEE/CVF Conference on Computer Vision and Pattern Recognition}, pages 22563--22575, 2023{\natexlab{b}}.

\bibitem[Chang et~al.(2022)Chang, Zhang, Jiang, Liu, and Freeman]{chang2022maskgit}
Huiwen Chang, Han Zhang, Lu Jiang, Ce Liu, and William~T Freeman.
\newblock Maskgit: Masked generative image transformer.
\newblock In \emph{Proceedings of the IEEE/CVF Conference on Computer Vision and Pattern Recognition}, pages 11315--11325, 2022.

\bibitem[Chang et~al.(2023)Chang, Zhang, Barber, Maschinot, Lezama, Jiang, Yang, Murphy, Freeman, Rubinstein, et~al.]{chang2023muse}
Huiwen Chang, Han Zhang, Jarred Barber, AJ Maschinot, Jose Lezama, Lu Jiang, Ming-Hsuan Yang, Kevin Murphy, William~T Freeman, Michael Rubinstein, et~al.
\newblock Muse: Text-to-image generation via masked generative transformers.
\newblock \emph{arXiv preprint arXiv:2301.00704}, 2023.

\bibitem[Chefer et~al.(2023)Chefer, Alaluf, Vinker, Wolf, and Cohen-Or]{chefer2023attend}
Hila Chefer, Yuval Alaluf, Yael Vinker, Lior Wolf, and Daniel Cohen-Or.
\newblock Attend-and-excite: Attention-based semantic guidance for text-to-image diffusion models.
\newblock In \emph{ACM Trans. Graph.}, 2023.

\bibitem[Chen et~al.(2024{\natexlab{a}})Chen, Zhang, Cun, Xia, Wang, Weng, and Shan]{chen2024videocrafter2}
Haoxin Chen, Yong Zhang, Xiaodong Cun, Menghan Xia, Xintao Wang, Chao Weng, and Ying Shan.
\newblock Videocrafter2: Overcoming data limitations for high-quality video diffusion models.
\newblock \emph{arXiv preprint arXiv:2401.09047}, 2024{\natexlab{a}}.

\bibitem[Chen et~al.(2024{\natexlab{b}})Chen, Laina, and Vedaldi]{chen2024training}
Minghao Chen, Iro Laina, and Andrea Vedaldi.
\newblock Training-free layout control with cross-attention guidance.
\newblock In \emph{WACV}, 2024{\natexlab{b}}.

\bibitem[Chen et~al.(2023)Chen, Liu, Yang, Yuan, You, Liu, and Yang]{chen2023reason}
Xiaohui Chen, Yongfei Liu, Yingxiang Yang, Jianbo Yuan, Quanzeng You, Li-Ping Liu, and Hongxia Yang.
\newblock Reason out your layout: Evoking the layout master from large language models for text-to-image synthesis.
\newblock \emph{arXiv preprint arXiv:2311.17126}, 2023.

\bibitem[Deng et~al.(2023)Deng, Gu, Zheng, Chen, Stevens, Wang, Sun, and Su]{deng2023mind2webgeneralistagentweb}
Xiang Deng, Yu Gu, Boyuan Zheng, Shijie Chen, Samuel Stevens, Boshi Wang, Huan Sun, and Yu Su.
\newblock Mind2web: Towards a generalist agent for the web, 2023.

\bibitem[Driess et~al.(2023)Driess, Xia, Sajjadi, Lynch, Chowdhery, Ichter, Wahid, Tompson, Vuong, Yu, Huang, Chebotar, Sermanet, Duckworth, Levine, Vanhoucke, Hausman, Toussaint, Greff, Zeng, Mordatch, and Florence]{driess2023palmeembodiedmultimodallanguage}
Danny Driess, Fei Xia, Mehdi S.~M. Sajjadi, Corey Lynch, Aakanksha Chowdhery, Brian Ichter, Ayzaan Wahid, Jonathan Tompson, Quan Vuong, Tianhe Yu, Wenlong Huang, Yevgen Chebotar, Pierre Sermanet, Daniel Duckworth, Sergey Levine, Vincent Vanhoucke, Karol Hausman, Marc Toussaint, Klaus Greff, Andy Zeng, Igor Mordatch, and Pete Florence.
\newblock Palm-e: An embodied multimodal language model, 2023.

\bibitem[Feng et~al.(2023)Feng, He, Fu, Jampani, Akula, Narayana, Basu, Wang, and Wang]{feng2022training}
Weixi Feng, Xuehai He, Tsu-Jui Fu, Varun Jampani, Arjun Akula, Pradyumna Narayana, Sugato Basu, Xin~Eric Wang, and William~Yang Wang.
\newblock Training-free structured diffusion guidance for compositional text-to-image synthesis.
\newblock In \emph{ICLR}, 2023.

\bibitem[Gani et~al.(2023)Gani, Bhat, Naseer, Khan, and Wonka]{gani2023llm}
Hanan Gani, Shariq~Farooq Bhat, Muzammal Naseer, Salman Khan, and Peter Wonka.
\newblock Llm blueprint: Enabling text-to-image generation with complex and detailed prompts.
\newblock \emph{arXiv preprint arXiv:2310.10640}, 2023.

\bibitem[Guo et~al.(2023)Guo, Yang, Rao, Wang, Qiao, Lin, and Dai]{guo2023animatediff}
Yuwei Guo, Ceyuan Yang, Anyi Rao, Yaohui Wang, Yu Qiao, Dahua Lin, and Bo Dai.
\newblock Animatediff: Animate your personalized text-to-image diffusion models without specific tuning.
\newblock \emph{arXiv preprint arXiv:2307.04725}, 2023.

\bibitem[He et~al.(2022)He, Yang, Zhang, Shan, and Chen]{he2022latent}
Yingqing He, Tianyu Yang, Yong Zhang, Ying Shan, and Qifeng Chen.
\newblock Latent video diffusion models for high-fidelity long video generation.
\newblock \emph{arXiv preprint arXiv:2211.13221}, 2022.

\bibitem[Ho et~al.(2020)Ho, Jain, and Abbeel]{ho2020denoisingdiffusionprobabilisticmodels}
Jonathan Ho, Ajay Jain, and Pieter Abbeel.
\newblock Denoising diffusion probabilistic models, 2020.

\bibitem[Ho et~al.(2022)Ho, Chan, Saharia, Whang, Gao, Gritsenko, Kingma, Poole, Norouzi, Fleet, et~al.]{ho2022imagen}
Jonathan Ho, William Chan, Chitwan Saharia, Jay Whang, Ruiqi Gao, Alexey Gritsenko, Diederik~P Kingma, Ben Poole, Mohammad Norouzi, David~J Fleet, et~al.
\newblock Imagen video: High definition video generation with diffusion models.
\newblock \emph{arXiv preprint arXiv:2210.02303}, 2022.

\bibitem[Hong et~al.(2024)Hong, Zhuge, Chen, Zheng, Cheng, Zhang, Wang, Wang, Yau, Lin, Zhou, Ran, Xiao, Wu, and Schmidhuber]{hong2024metagptmetaprogrammingmultiagent}
Sirui Hong, Mingchen Zhuge, Jiaqi Chen, Xiawu Zheng, Yuheng Cheng, Ceyao Zhang, Jinlin Wang, Zili Wang, Steven Ka~Shing Yau, Zijuan Lin, Liyang Zhou, Chenyu Ran, Lingfeng Xiao, Chenglin Wu, and Jürgen Schmidhuber.
\newblock Metagpt: Meta programming for a multi-agent collaborative framework, 2024.

\bibitem[Hong et~al.(2022)Hong, Ding, Zheng, Liu, and Tang]{hong2022cogvideo}
Wenyi Hong, Ming Ding, Wendi Zheng, Xinghan Liu, and Jie Tang.
\newblock Cogvideo: Large-scale pretraining for text-to-video generation via transformers.
\newblock \emph{arXiv preprint arXiv:2205.15868}, 2022.

\bibitem[hpcaitech(2024)]{opensora}
hpcaitech.
\newblock Open-sora: Democratizing efficient video production for all, 2024.

\bibitem[Huang et~al.(2024)Huang, Sun, Xie, Li, and Liu]{huang2024t2i}
Kaiyi Huang, Kaiyue Sun, Enze Xie, Zhenguo Li, and Xihui Liu.
\newblock T2i-compbench: A comprehensive benchmark for open-world compositional text-to-image generation.
\newblock \emph{NeurIPS}, 2024.

\bibitem[Khachatryan et~al.(2023)Khachatryan, Movsisyan, Tadevosyan, Henschel, Wang, Navasardyan, and Shi]{khachatryan2023text2video}
Levon Khachatryan, Andranik Movsisyan, Vahram Tadevosyan, Roberto Henschel, Zhangyang Wang, Shant Navasardyan, and Humphrey Shi.
\newblock Text2video-zero: Text-to-image diffusion models are zero-shot video generators.
\newblock In \emph{Proceedings of the IEEE/CVF International Conference on Computer Vision}, pages 15954--15964, 2023.

\bibitem[Kim et~al.(2023)Kim, Lee, Kim, Ha, and Zhu]{kim2023dense}
Yunji Kim, Jiyoung Lee, Jin-Hwa Kim, Jung-Woo Ha, and Jun-Yan Zhu.
\newblock Dense text-to-image generation with attention modulation.
\newblock In \emph{ICCV}, 2023.

\bibitem[Kondratyuk et~al.(2023)Kondratyuk, Yu, Gu, Lezama, Huang, Hornung, Adam, Akbari, Alon, Birodkar, et~al.]{kondratyuk2023videopoet}
Dan Kondratyuk, Lijun Yu, Xiuye Gu, Jos{\'e} Lezama, Jonathan Huang, Rachel Hornung, Hartwig Adam, Hassan Akbari, Yair Alon, Vighnesh Birodkar, et~al.
\newblock Videopoet: A large language model for zero-shot video generation.
\newblock \emph{arXiv preprint arXiv:2312.14125}, 2023.

\bibitem[Lab and etc.(2024)]{pku_yuan_lab_and_tuzhan_ai_etc_2024_10948109}
PKU-Yuan Lab and Tuzhan~AI etc.
\newblock Open-sora-plan, 2024.

\bibitem[Li et~al.(2023)Li, Liu, Wu, Mu, Yang, Gao, Li, and Lee]{li2023gligen}
Yuheng Li, Haotian Liu, Qingyang Wu, Fangzhou Mu, Jianwei Yang, Jianfeng Gao, Chunyuan Li, and Yong~Jae Lee.
\newblock Gligen: Open-set grounded text-to-image generation.
\newblock In \emph{ICCV}, 2023.

\bibitem[Li et~al.(2022)Li, Min, Li, and Xu]{li2022stylet2i}
Zhiheng Li, Martin~Renqiang Min, Kai Li, and Chenliang Xu.
\newblock Stylet2i: Toward compositional and high-fidelity text-to-image synthesis.
\newblock In \emph{CVPR}, 2022.

\bibitem[Lian et~al.(2023)Lian, Shi, Yala, Darrell, and Li]{lian2023llm}
Long Lian, Baifeng Shi, Adam Yala, Trevor Darrell, and Boyi Li.
\newblock Llm-grounded video diffusion models.
\newblock \emph{arXiv preprint arXiv:2309.17444}, 2023.

\bibitem[Lin et~al.(2024)Lin, Zala, Cho, and Bansal]{Lin2023VideoDirectorGPT}
Han Lin, Abhay Zala, Jaemin Cho, and Mohit Bansal.
\newblock Videodirectorgpt: Consistent multi-scene video generation via llm-guided planning.
\newblock In \emph{COLM}, 2024.

\bibitem[Liu et~al.(2022)Liu, Li, Du, Torralba, and Tenenbaum]{liu2022compositional}
Nan Liu, Shuang Li, Yilun Du, Antonio Torralba, and Joshua~B Tenenbaum.
\newblock Compositional visual generation with composable diffusion models.
\newblock In \emph{ECCV}, 2022.

\bibitem[Liu et~al.(2024)Liu, Hu, Wang, Kawaguchi, and Yao]{liu2024referee}
Xuantong Liu, Tianyang Hu, Wenjia Wang, Kenji Kawaguchi, and Yuan Yao.
\newblock Referee can play: An alternative approach to conditional generation via model inversion.
\newblock \emph{arXiv preprint arXiv:2402.16305}, 2024.

\bibitem[Luo et~al.(2023)Luo, Chen, Zhang, Huang, Wang, Shen, Zhao, Zhou, and Tan]{luo2023videofusion}
Zhengxiong Luo, Dayou Chen, Yingya Zhang, Yan Huang, Liang Wang, Yujun Shen, Deli Zhao, Jingren Zhou, and Tieniu Tan.
\newblock Videofusion: Decomposed diffusion models for high-quality video generation.
\newblock In \emph{Proceedings of the IEEE/CVF Conference on Computer Vision and Pattern Recognition}, pages 10209--10218, 2023.

\bibitem[Ma et~al.(2024)Ma, Wang, Jia, Chen, Liu, Li, Chen, and Qiao]{ma2024latte}
Xin Ma, Yaohui Wang, Gengyun Jia, Xinyuan Chen, Ziwei Liu, Yuan-Fang Li, Cunjian Chen, and Yu Qiao.
\newblock Latte: Latent diffusion transformer for video generation.
\newblock \emph{arXiv preprint arXiv:2401.03048}, 2024.

\bibitem[Meral et~al.(2023)Meral, Simsar, Tombari, and Yanardag]{meral2023conform}
Tuna Han~Salih Meral, Enis Simsar, Federico Tombari, and Pinar Yanardag.
\newblock Conform: Contrast is all you need for high-fidelity text-to-image diffusion models.
\newblock \emph{arXiv preprint arXiv:2312.06059}, 2023.

\bibitem[OpenAI(2024)]{gpt-4o}
OpenAI.
\newblock {Hello GPT-4o}.
\newblock \url{https://openai.com/index/hello-gpt-4o/}, 2024.
\newblock Accessed: 2024-11-14.

\bibitem[Park et~al.(2021)Park, Azadi, Liu, Darrell, and Rohrbach]{park2021benchmark}
Dong~Huk Park, Samaneh Azadi, Xihui Liu, Trevor Darrell, and Anna Rohrbach.
\newblock Benchmark for compositional text-to-image synthesis.
\newblock In \emph{NeurIPS}, 2021.

\bibitem[Park et~al.(2023)Park, O'Brien, Cai, Morris, Liang, and Bernstein]{park2023generativeagentsinteractivesimulacra}
Joon~Sung Park, Joseph~C. O'Brien, Carrie~J. Cai, Meredith~Ringel Morris, Percy Liang, and Michael~S. Bernstein.
\newblock Generative agents: Interactive simulacra of human behavior, 2023.

\bibitem[Patel et~al.(2023)Patel, Kim, Cheng, Baral, and Yang]{patel2023eclipse}
Maitreya Patel, Changhoon Kim, Sheng Cheng, Chitta Baral, and Yezhou Yang.
\newblock Eclipse: A resource-efficient text-to-image prior for image generations.
\newblock \emph{arXiv preprint arXiv:2312.04655}, 2023.

\bibitem[Qian et~al.(2024)Qian, Liu, Liu, Chen, Dang, Li, Yang, Chen, Su, Cong, Xu, Li, Liu, and Sun]{qian2024chatdevcommunicativeagentssoftware}
Chen Qian, Wei Liu, Hongzhang Liu, Nuo Chen, Yufan Dang, Jiahao Li, Cheng Yang, Weize Chen, Yusheng Su, Xin Cong, Juyuan Xu, Dahai Li, Zhiyuan Liu, and Maosong Sun.
\newblock Chatdev: Communicative agents for software development, 2024.

\bibitem[Rassin et~al.(2024)Rassin, Hirsch, Glickman, Ravfogel, Goldberg, and Chechik]{rassin2024linguistic}
Royi Rassin, Eran Hirsch, Daniel Glickman, Shauli Ravfogel, Yoav Goldberg, and Gal Chechik.
\newblock Linguistic binding in diffusion models: Enhancing attribute correspondence through attention map alignment.
\newblock \emph{NeurIPS}, 2024.

\bibitem[{Runway AI}(2024)]{gen3}
{Runway AI}.
\newblock Gen-3.
\newblock \url{https://runwayml.com/blog/introducing-gen-3-alpha/}, 2024.
\newblock Accessed: 2024-11-14.

\bibitem[Singer et~al.(2022)Singer, Polyak, Hayes, Yin, An, Zhang, Hu, Yang, Ashual, Gafni, et~al.]{singer2022make}
Uriel Singer, Adam Polyak, Thomas Hayes, Xi Yin, Jie An, Songyang Zhang, Qiyuan Hu, Harry Yang, Oron Ashual, Oran Gafni, et~al.
\newblock Make-a-video: Text-to-video generation without text-video data.
\newblock \emph{arXiv preprint arXiv:2209.14792}, 2022.

\bibitem[Sohl-Dickstein et~al.(2015)Sohl-Dickstein, Weiss, Maheswaranathan, and Ganguli]{sohldickstein2015deepunsupervisedlearningusing}
Jascha Sohl-Dickstein, Eric~A. Weiss, Niru Maheswaranathan, and Surya Ganguli.
\newblock Deep unsupervised learning using nonequilibrium thermodynamics, 2015.

\bibitem[Song et~al.(2021)Song, Sohl-Dickstein, Kingma, Kumar, Ermon, and Poole]{song2021scorebasedgenerativemodelingstochastic}
Yang Song, Jascha Sohl-Dickstein, Diederik~P. Kingma, Abhishek Kumar, Stefano Ermon, and Ben Poole.
\newblock Score-based generative modeling through stochastic differential equations, 2021.

\bibitem[Sun et~al.(2024{\natexlab{a}})Sun, Huang, Liu, Wu, Xu, Li, and Liu]{sun2024t2vcompbench}
Kaiyue Sun, Kaiyi Huang, Xian Liu, Yue Wu, Zihan Xu, Zhenguo Li, and Xihui Liu.
\newblock T2v-compbench: A comprehensive benchmark for compositional text-to-video generation, 2024{\natexlab{a}}.

\bibitem[Sun et~al.(2024{\natexlab{b}})Sun, Chen, Xu, Cheng, Ma, Yin, Wang, Han, Zhu, Yuan, Guo, Qiu, Yin, Li, Yuan, Kong, Li, and Wu]{sun2024surveyneuralcodeintelligence}
Qiushi Sun, Zhirui Chen, Fangzhi Xu, Kanzhi Cheng, Chang Ma, Zhangyue Yin, Jianing Wang, Chengcheng Han, Renyu Zhu, Shuai Yuan, Qipeng Guo, Xipeng Qiu, Pengcheng Yin, Xiaoli Li, Fei Yuan, Lingpeng Kong, Xiang Li, and Zhiyong Wu.
\newblock A survey of neural code intelligence: Paradigms, advances and beyond, 2024{\natexlab{b}}.

\bibitem[Sun et~al.(2024{\natexlab{c}})Sun, Yin, Li, Wu, Qiu, and Kong]{sun2024corexpushingboundariescomplex}
Qiushi Sun, Zhangyue Yin, Xiang Li, Zhiyong Wu, Xipeng Qiu, and Lingpeng Kong.
\newblock Corex: Pushing the boundaries of complex reasoning through multi-model collaboration, 2024{\natexlab{c}}.

\bibitem[Taghipour et~al.(2024)Taghipour, Ghahremani, Bennamoun, Rekavandi, Laga, and Boussaid]{taghipour2024box}
Ashkan Taghipour, Morteza Ghahremani, Mohammed Bennamoun, Aref~Miri Rekavandi, Hamid Laga, and Farid Boussaid.
\newblock Box it to bind it: Unified layout control and attribute binding in t2i diffusion models.
\newblock \emph{arXiv preprint arXiv:2402.17910}, 2024.

\bibitem[Tang et~al.(2024)Tang, Jin, Zhu, Yuan, Zhang, Zhou, Qu, Zhao, Tang, Zhang, Cohan, Lu, and Gerstein]{tang2024prioritizingsafeguardingautonomyrisks}
Xiangru Tang, Qiao Jin, Kunlun Zhu, Tongxin Yuan, Yichi Zhang, Wangchunshu Zhou, Meng Qu, Yilun Zhao, Jian Tang, Zhuosheng Zhang, Arman Cohan, Zhiyong Lu, and Mark Gerstein.
\newblock Prioritizing safeguarding over autonomy: Risks of llm agents for science, 2024.

\bibitem[Tian et~al.(2024)Tian, Yang, Yang, Gao, Deng, Chen, Wang, Yu, Tao, Wan, Zhang, and Cui]{videoteris}
Ye Tian, Ling Yang, Haotian Yang, Yuan Gao, Yufan Deng, Jingmin Chen, Xintao Wang, Zhaochen Yu, Xin Tao, Pengfei Wan, Di Zhang, and Bin Cui.
\newblock Videotetris: Towards compositional text-to-video generation, 2024.

\bibitem[Villegas et~al.(2022)Villegas, Babaeizadeh, Kindermans, Moraldo, Zhang, Saffar, Castro, Kunze, and Erhan]{villegas2022phenaki}
Ruben Villegas, Mohammad Babaeizadeh, Pieter-Jan Kindermans, Hernan Moraldo, Han Zhang, Mohammad~Taghi Saffar, Santiago Castro, Julius Kunze, and Dumitru Erhan.
\newblock Phenaki: Variable length video generation from open domain textual descriptions.
\newblock In \emph{International Conference on Learning Representations}, 2022.

\bibitem[Wang et~al.(2023{\natexlab{a}})Wang, Xie, Jiang, Mandlekar, Xiao, Zhu, Fan, and Anandkumar]{wang2023voyageropenendedembodiedagent}
Guanzhi Wang, Yuqi Xie, Yunfan Jiang, Ajay Mandlekar, Chaowei Xiao, Yuke Zhu, Linxi Fan, and Anima Anandkumar.
\newblock Voyager: An open-ended embodied agent with large language models, 2023{\natexlab{a}}.

\bibitem[Wang et~al.(2023{\natexlab{b}})Wang, Yuan, Chen, Zhang, Wang, and Zhang]{wang2023modelscope}
Jiuniu Wang, Hangjie Yuan, Dayou Chen, Yingya Zhang, Xiang Wang, and Shiwei Zhang.
\newblock Modelscope text-to-video technical report.
\newblock \emph{arXiv preprint arXiv:2308.06571}, 2023{\natexlab{b}}.

\bibitem[Wang et~al.(2023{\natexlab{c}})Wang, Chen, Chen, Ma, Lu, and Lin]{wang2023compositional}
Ruichen Wang, Zekang Chen, Chen Chen, Jian Ma, Haonan Lu, and Xiaodong Lin.
\newblock Compositional text-to-image synthesis with attention map control of diffusion models.
\newblock \emph{arXiv preprint arXiv:2305.13921}, 2023{\natexlab{c}}.

\bibitem[Wang et~al.(2024{\natexlab{a}})Wang, Li, Song, Xu, Tang, Zhuge, Pan, Song, Li, Singh, Tran, Li, Ma, Zheng, Qian, Shao, Muennighoff, Zhang, Hui, Lin, Brennan, Peng, Ji, and Neubig]{wang2024openhandsopenplatformai}
Xingyao Wang, Boxuan Li, Yufan Song, Frank~F. Xu, Xiangru Tang, Mingchen Zhuge, Jiayi Pan, Yueqi Song, Bowen Li, Jaskirat Singh, Hoang~H. Tran, Fuqiang Li, Ren Ma, Mingzhang Zheng, Bill Qian, Yanjun Shao, Niklas Muennighoff, Yizhe Zhang, Binyuan Hui, Junyang Lin, Robert Brennan, Hao Peng, Heng Ji, and Graham Neubig.
\newblock Openhands: An open platform for ai software developers as generalist agents, 2024{\natexlab{a}}.

\bibitem[Wang et~al.(2024{\natexlab{b}})Wang, Li, Li, and Liu]{wang2024genartistmultimodalllmagent}
Zhenyu Wang, Aoxue Li, Zhenguo Li, and Xihui Liu.
\newblock Genartist: Multimodal llm as an agent for unified image generation and editing, 2024{\natexlab{b}}.

\bibitem[Wang et~al.(2024{\natexlab{c}})Wang, Xie, Li, Wang, Liu, and Li]{wang2024divide}
Zhenyu Wang, Enze Xie, Aoxue Li, Zhongdao Wang, Xihui Liu, and Zhenguo Li.
\newblock Divide and conquer: Language models can plan and self-correct for compositional text-to-image generation.
\newblock \emph{arXiv preprint arXiv:2401.15688}, 2024{\natexlab{c}}.

\bibitem[Wu et~al.(2021)Wu, Huang, Zhang, Li, Ji, Yang, Sapiro, and Duan]{wu2021godiva}
Chenfei Wu, Lun Huang, Qianxi Zhang, Binyang Li, Lei Ji, Fan Yang, Guillermo Sapiro, and Nan Duan.
\newblock Godiva: Generating open-domain videos from natural descriptions.
\newblock \emph{arXiv preprint arXiv:2104.14806}, 2021.

\bibitem[Wu et~al.(2022)Wu, Liang, Ji, Yang, Fang, Jiang, and Duan]{wu2022nuwa}
Chenfei Wu, Jian Liang, Lei Ji, Fan Yang, Yuejian Fang, Daxin Jiang, and Nan Duan.
\newblock N{\"u}wa: Visual synthesis pre-training for neural visual world creation.
\newblock In \emph{European conference on computer vision}, pages 720--736. Springer, 2022.

\bibitem[Wu et~al.(2023{\natexlab{a}})Wu, Bansal, Zhang, Wu, Li, Zhu, Jiang, Zhang, Zhang, Liu, Awadallah, White, Burger, and Wang]{wu2023autogenenablingnextgenllm}
Qingyun Wu, Gagan Bansal, Jieyu Zhang, Yiran Wu, Beibin Li, Erkang Zhu, Li Jiang, Xiaoyun Zhang, Shaokun Zhang, Jiale Liu, Ahmed~Hassan Awadallah, Ryen~W White, Doug Burger, and Chi Wang.
\newblock Autogen: Enabling next-gen llm applications via multi-agent conversation, 2023{\natexlab{a}}.

\bibitem[Wu et~al.(2023{\natexlab{b}})Wu, Liu, Zhao, Bui, Lin, Zhang, and Chang]{wu2023harnessing}
Qiucheng Wu, Yujian Liu, Handong Zhao, Trung Bui, Zhe Lin, Yang Zhang, and Shiyu Chang.
\newblock Harnessing the spatial-temporal attention of diffusion models for high-fidelity text-to-image synthesis.
\newblock In \emph{ICCV}, 2023{\natexlab{b}}.

\bibitem[Wu et~al.(2023{\natexlab{c}})Wu, Lian, Gonzalez, Li, and Darrell]{wu2023selfcorrectingllmcontrolleddiffusionmodels}
Tsung-Han Wu, Long Lian, Joseph~E. Gonzalez, Boyi Li, and Trevor Darrell.
\newblock Self-correcting llm-controlled diffusion models, 2023{\natexlab{c}}.

\bibitem[Yang et~al.(2024{\natexlab{a}})Yang, Yu, Meng, Xu, Ermon, and Cui]{yang2024mastering}
Ling Yang, Zhaochen Yu, Chenlin Meng, Minkai Xu, Stefano Ermon, and Bin Cui.
\newblock Mastering text-to-image diffusion: Recaptioning, planning, and generating with multimodal llms.
\newblock \emph{arXiv preprint arXiv:2401.11708}, 2024{\natexlab{a}}.

\bibitem[Yang and Wang(2024)]{yang2024vico}
Xingyi Yang and Xinchao Wang.
\newblock Compositional video generation as flow equalization, 2024.

\bibitem[Yang et~al.(2024{\natexlab{b}})Yang, Teng, Zheng, Ding, Huang, Xu, Yang, Hong, Zhang, Feng, et~al.]{yang2024cogvideox}
Zhuoyi Yang, Jiayan Teng, Wendi Zheng, Ming Ding, Shiyu Huang, Jiazheng Xu, Yuanming Yang, Wenyi Hong, Xiaohan Zhang, Guanyu Feng, et~al.
\newblock Cogvideox: Text-to-video diffusion models with an expert transformer.
\newblock \emph{arXiv preprint arXiv:2408.06072}, 2024{\natexlab{b}}.

\bibitem[Yao et~al.(2023)Yao, Chen, Yang, and Narasimhan]{yao2023webshopscalablerealworldweb}
Shunyu Yao, Howard Chen, John Yang, and Karthik Narasimhan.
\newblock Webshop: Towards scalable real-world web interaction with grounded language agents, 2023.

\bibitem[Yu et~al.(2023{\natexlab{a}})Yu, Cheng, Sohn, Lezama, Zhang, Chang, Hauptmann, Yang, Hao, Essa, et~al.]{yu2023magvit}
Lijun Yu, Yong Cheng, Kihyuk Sohn, Jos{\'e} Lezama, Han Zhang, Huiwen Chang, Alexander~G Hauptmann, Ming-Hsuan Yang, Yuan Hao, Irfan Essa, et~al.
\newblock Magvit: Masked generative video transformer.
\newblock In \emph{Proceedings of the IEEE/CVF Conference on Computer Vision and Pattern Recognition}, pages 10459--10469, 2023{\natexlab{a}}.

\bibitem[Yu et~al.(2023{\natexlab{b}})Yu, Lezama, Gundavarapu, Versari, Sohn, Minnen, Cheng, Gupta, Gu, Hauptmann, et~al.]{yu2023language}
Lijun Yu, Jos{\'e} Lezama, Nitesh~B Gundavarapu, Luca Versari, Kihyuk Sohn, David Minnen, Yong Cheng, Agrim Gupta, Xiuye Gu, Alexander~G Hauptmann, et~al.
\newblock Language model beats diffusion--tokenizer is key to visual generation.
\newblock \emph{arXiv preprint arXiv:2310.05737}, 2023{\natexlab{b}}.

\bibitem[Yuan et~al.(2024{\natexlab{a}})Yuan, Huang, Shi, Xu, Zhu, Lin, Cheng, Yuan, and Luo]{yuan2024magictime}
Shenghai Yuan, Jinfa Huang, Yujun Shi, Yongqi Xu, Ruijie Zhu, Bin Lin, Xinhua Cheng, Li Yuan, and Jiebo Luo.
\newblock Magictime: Time-lapse video generation models as metamorphic simulators.
\newblock \emph{arXiv preprint arXiv:2404.05014}, 2024{\natexlab{a}}.

\bibitem[Yuan et~al.(2024{\natexlab{b}})Yuan, Liu, Cao, Sun, Jia, Chen, Li, Lin, Yuan, He, Wang, Ye, and Sun]{yuan2024moraenablinggeneralistvideo}
Zhengqing Yuan, Yixin Liu, Yihan Cao, Weixiang Sun, Haolong Jia, Ruoxi Chen, Zhaoxu Li, Bin Lin, Li Yuan, Lifang He, Chi Wang, Yanfang Ye, and Lichao Sun.
\newblock Mora: Enabling generalist video generation via a multi-agent framework, 2024{\natexlab{b}}.

\bibitem[Zhang et~al.(2023)Zhang, Wu, Liu, Zhao, Ran, Gu, Gao, and Shou]{zhang2023show}
David~Junhao Zhang, Jay~Zhangjie Wu, Jia-Wei Liu, Rui Zhao, Lingmin Ran, Yuchao Gu, Difei Gao, and Mike~Zheng Shou.
\newblock Show-1: Marrying pixel and latent diffusion models for text-to-video generation.
\newblock \emph{arXiv preprint arXiv:2309.15818}, 2023.

\bibitem[Zhou et~al.(2022)Zhou, Wang, Yan, Lv, Zhu, and Feng]{zhou2022magicvideo}
Daquan Zhou, Weimin Wang, Hanshu Yan, Weiwei Lv, Yizhe Zhu, and Jiashi Feng.
\newblock Magicvideo: Efficient video generation with latent diffusion models.
\newblock \emph{arXiv preprint arXiv:2211.11018}, 2022.

\end{thebibliography}
}

\clearpage
\setcounter{page}{1}
\maketitlesupplementary
\appendix

\renewcommand\thefigure{A\arabic{figure}}
\renewcommand\thetable{A\arabic{table}}  
\renewcommand\theequation{A\arabic{equation}}

\section{Framework Details}
\label{app: generation_intensity}
\textbf{Details in the \generation{} Stage.}
In the \generation{} stage, the video is generated based on the structured layout predicted in the \design{} stage. The video is then progressively refined through the \redesign{}-\generation{} loop by adjusting multiple types of guidance including structured layout, guidance scale, and text prompt dynamically.

Following the approach in LVD~\cite{lian2023llm}, we utilize the scene layouts across frames predicted by the LLM in the \design{} stage to direct the initial video generation. During the denoising process, the generation model integrates the information from the text prompt into the latent features via cross-attention layers. %
To ensure an object appears within its designated bounding box, LVD~\cite{lian2023llm} designs an energy function $\mathcal{L}$ that enforces this constraint.
By applying gradient descent of this energy function on partially denoised frames, we can gradually align the video output with the compositional layout specified by the prompt.

Specifically, given the text prompt $\mathcal{P}$, we extract all object tokens (\eg, nouns) as $\mathcal{O} = \{o_1, ..., o_k\}$ from  $\mathcal{P}$. For each object token in $\mathcal{O}$, our goal is to encourage the values of $A_t^o$ within the designated bounding box region to be high, where $A_t^o$ denotes the cross-attention map from the latent layers to the object token $o$ at timestep $t$. The energy function is defined as: 
\begin{equation}
    \begin{split}
      \mathcal{L} &=  \sum_{o \in O} \mathcal{L}_o \\
      \mathcal{L}_o &=  -\beta \cdot \texttt{Topk}(A_t^o \cdot M_t^o) \\
                    &\quad + \texttt{Topk}\big(A_t^o \cdot (1 - M_t^o)\big),      
    \end{split}
\end{equation}
where $M_t^o$ is a mask indicating the designated bounding box region (ones inside and zeros outside the bounding box) for the object token $o$ at timestep $t$. $\texttt{Topk}$ computes the average of top-$k$ values in a matrix. This loss function encourages high attention values within the bounding box region. $\beta$ denotes the guidance scale.

After computing $\mathcal{L}$, we update the latent feature $z_t$ by descending in the direction of its gradient:
\begin{equation}~\label{eq:latent_update}
    z'_t \gets z_t - \alpha_t \cdot  \nabla_{z_t} \mathcal{L},
\end{equation}
where $\alpha_t$ is a scalar defining the step size of the update.
This procedure is applied for a subset of denoising timesteps $t = T,T-1,\dots,t_{end}$. 

Note that different from LVD~\cite{lian2023llm} that uses a fix input text prompt, and predefined guidance scale, our \redesign{} stage dynamically adjusts the text prompt and guidance scale by multi-agent collaboration.

\noindent
\textbf{Formulation of Agents in the \redesign{} Stage.}
We assume that the iterative refinement loop at iteration $i$ can be denoted as $(\rho_i, \epsilon_i, \mathcal{V}_i)$, where $\rho$ denotes \texttt{Reasoning} texts in the form of flexible natural language, $\epsilon$ denotes \texttt{Execution} texts in structured forms, $\mathcal{V}$ denotes the \texttt{Video}. 
To be more concrete, \texttt{Reasoning} denotes the MLLM-output text for video understanding, verification, and correction, while \texttt{Execution} means to translate the reasoning results into a more structured format.
$\epsilon$ can be further expressed as $(\epsilon^c, \epsilon^s)$, where $\epsilon^c$ denotes the appropriate selected \refineagent{}, and $\epsilon^s$ denotes the structured outputs. 
Let $\pi_{\rm veri}$, $\pi_{\rm sugg}$, $\pi_{\rm corr}$, and $\pi_{\rm output}$ be the \alignmentagent{}, \suggagent{}, \refineagent{}, and \structuredagent{} separately.
$\rho_i'$ and $\rho_i''$ represent refined reasoning steps that build upon $\rho_i$ through successive agents $\pi_{\rm sugg}$ and $\pi_{\rm corr}$, integrating additional execution choices and previous structured outputs. 
$\mathcal{S}$ represents the prompts containing target task information and role allocation requirements. 
Then, the responsibilities and workflows of each agent are defined as follows: 
\begin{align}
    \label{eq:veri}
    \rho_i &= \pi_{\rm veri}(\mathcal{S}, \mathcal{V}_i, \mathcal{P}), \\
        \label{eq:sugg}
    \rho_i', \epsilon_i^c &= \pi_{\rm sugg}(\mathcal{S}, \mathcal{V}_i,  \rho_i), \\
    \label{eq:corr}
    \rho_i'' &= \pi_{\rm corr}(\mathcal{S}, \mathcal{V}_i,  \rho_i', \epsilon_i^c, \epsilon_{i-1}^s), \\
    \label{eq:structure}
    \epsilon_i^s &= \pi_{\rm output}(\mathcal{S}, \mathcal{V}_i, \rho_i'').
\end{align}
Note that $\pi_{\rm corr}$ uses the structured output from the previous iteration ($\epsilon_{i-1}^s$), as a reference to evaluate the current suggestion ($\epsilon_i^c$) and guide decisions on whether and how to revise it. By analyzing $\epsilon_{i-1}^s$ and $\epsilon_i^c$, $\pi_{\rm corr}$ can identify discrepancies, and unresolved issues. $\pi_{\rm corr}$ compares the structured output $\epsilon_{i-1}^s$ with the new suggestion from $\pi_{\rm sugg}$ to reason about the need for revisions. This process involves evaluating whether the layout needs adjustments (\textit{e.g.}, structural changes) or whether guidance scales should be emphasized to better align with the desired outcomes. The full examples are illustrated in~\Cref{tab:app_full_example3} and ~\Cref{tab:app_full_example2}.

\section{Additional Experimental Results}
\subsection{Implementation Details.}
\label{app: implementation}
We apply our~\name~on VideoCrafter2~\cite{chen2024videocrafter2} as the backbone for the \generation{} stage to generate videos with 65 frames, 512$\times$512 resolution. We find that VideoCrafter2 already works well, and our framework is compatible with other models too. Using more advanced models could potentially enhance performances, which we leave for future exploration.
We set 1.0 as the initialized guidance scale, and 0.05 as the incremental step.

\begin{figure*}[ht]
	\centering
	\begin{minipage}[c]{0.45\textwidth}
		\centering
		\includegraphics[width=\textwidth]{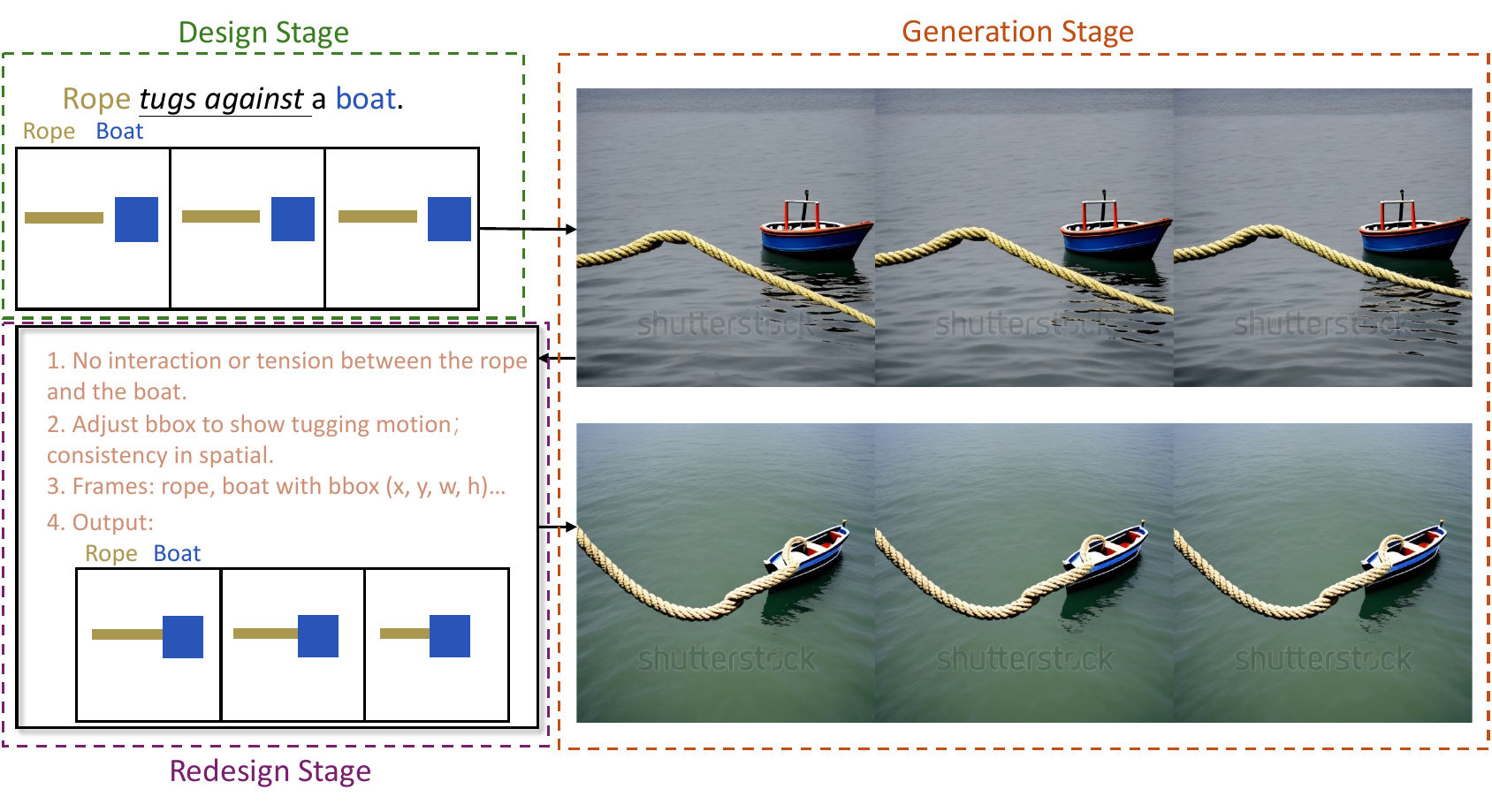}
		\subcaption{\textbf{Visualization} of multi-agent collaboration. Initial generation lacks ``tug'' motion between the rope and boat; \redesign{} agents adjust spatial alignment and visual tension, leading to a final video that aligns with the prompt’s interaction requirements.}
		\label{fig:qualitative_1}
	\end{minipage} 
	\begin{minipage}[c]{0.45\textwidth}
		\centering
		\includegraphics[width=\textwidth]{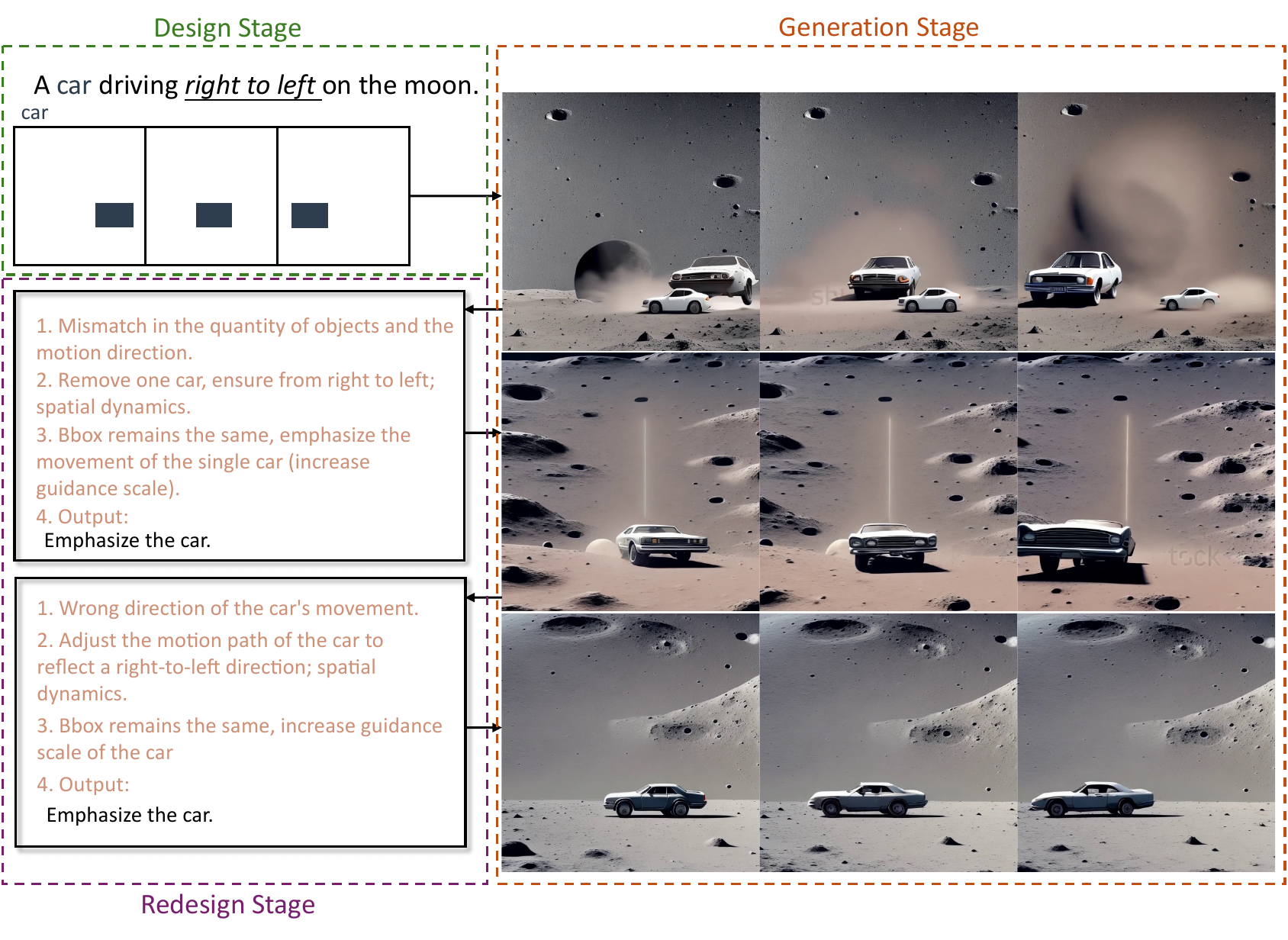}
		\subcaption{\textbf{Visualization} of the iterative refinement in correcting object quantity and motion direction. The \redesign{} agents adjust guidance scale and alignment over successive iterations, progressively enhancing adherence to the prompt.}
		\label{fig:qualitative_3}
	\end{minipage}
	\caption{\textbf{Visualization} of the multi-agent collaboration.}
	\label{fig:app_qualitative}
\end{figure*}

\begin{figure*}[ht]
	\centering
	\begin{minipage}[c]{0.32\textwidth}
		\centering
		\includegraphics[width=\textwidth]{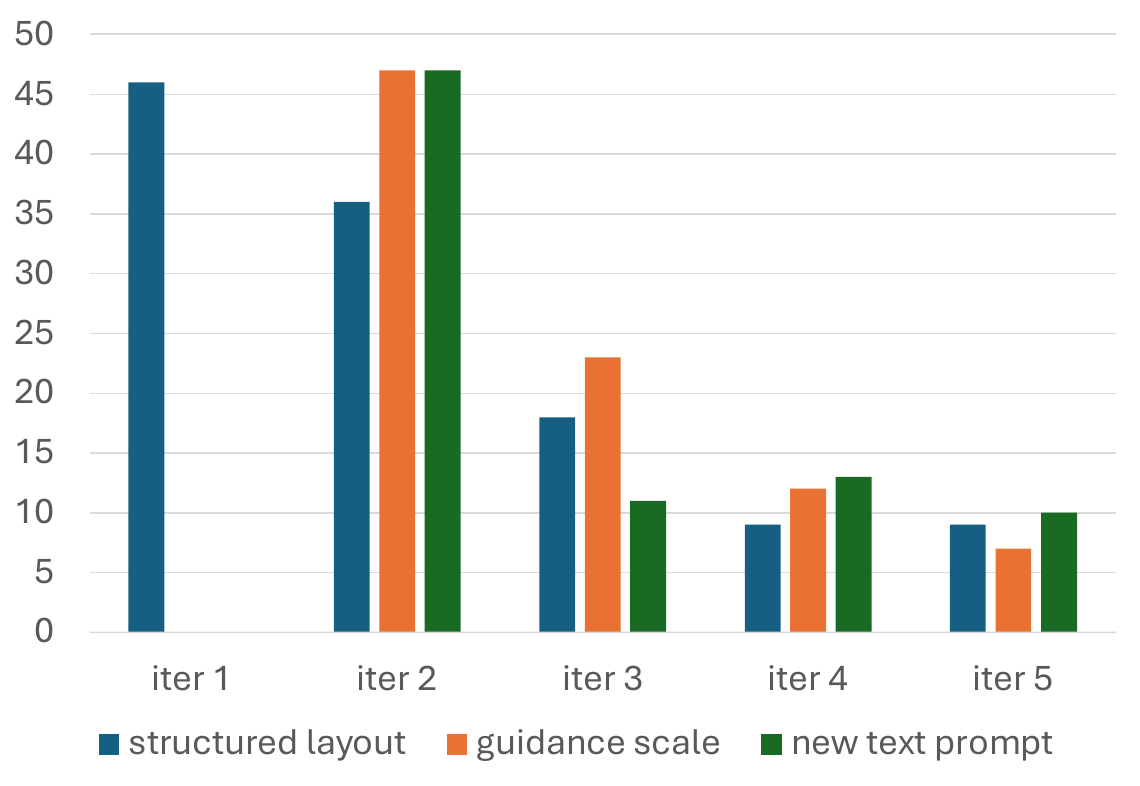}
		\subcaption{The number of corrections.}
		\label{fig:multi_guidance_num}
	\end{minipage} 
	\begin{minipage}[c]{0.32\textwidth}
		\centering
		\includegraphics[width=\textwidth]{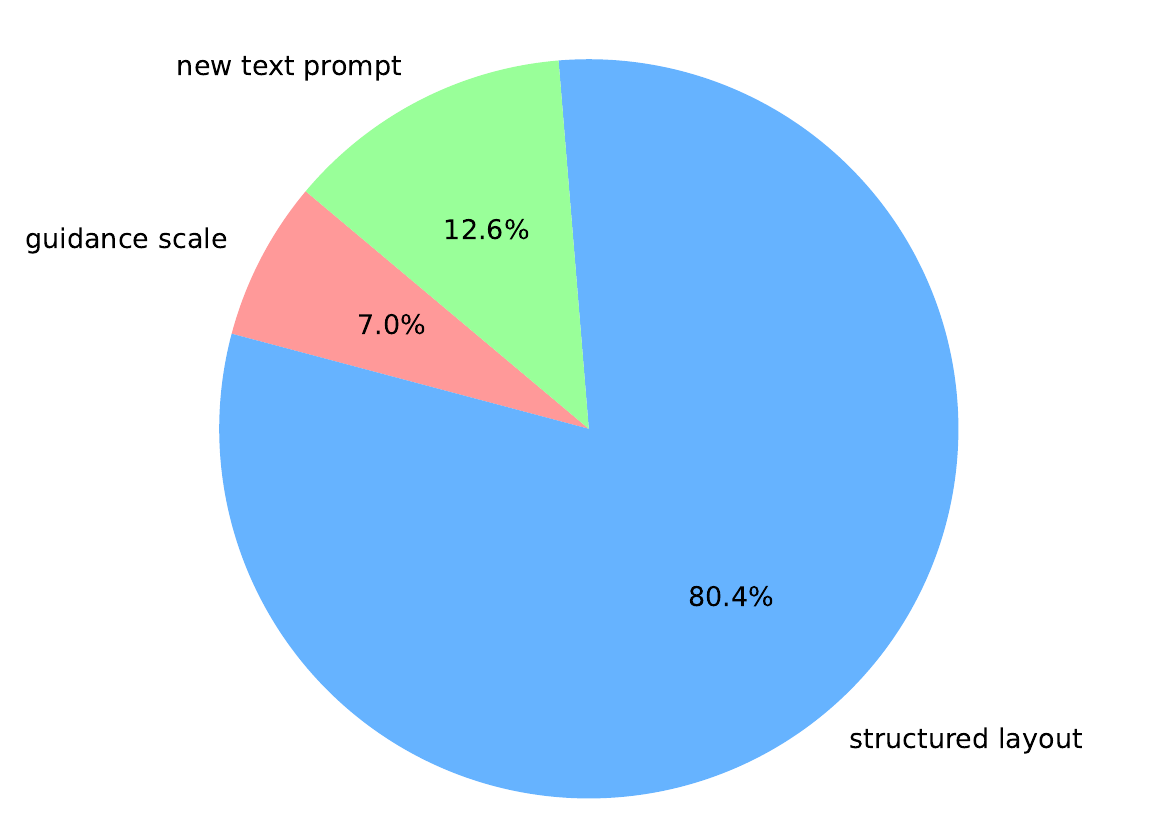}
		\subcaption{
        The contribution (\%) of different guidance types to the video scores with \design{} and \redesign{} stages.}
		\label{fig:multi_guidance_w_iter1}
	\end{minipage} 
	\begin{minipage}[c]{0.32\textwidth}
		\centering
		\includegraphics[width=\textwidth]{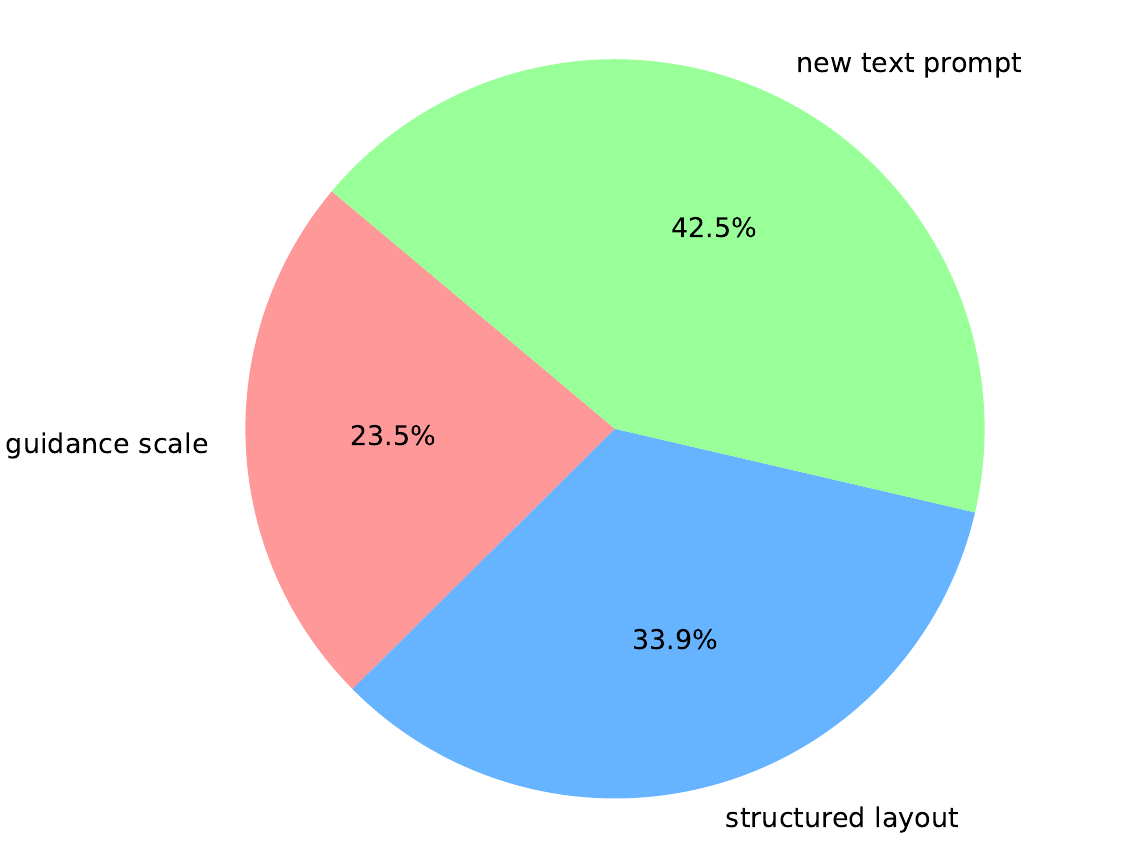}
		\subcaption{        The contribution (\%) of different guidance types to the video scores with only the \redesign{} stage.}
		\label{fig:multi_guidance_wo_iter1}
	\end{minipage}
	\caption{Illustration of the number of corrections and contributions (\%) in T2V-CompBench of different guidance types: structured layout, guidance scale, and new text prompt.}
	\label{fig:multi_guidance}
\end{figure*}

\subsection{Qualitative Comparisons}
\label{app: qualitative_compare}
\begin{figure*}[h]   %
  \centering
   \includegraphics[width=\linewidth]{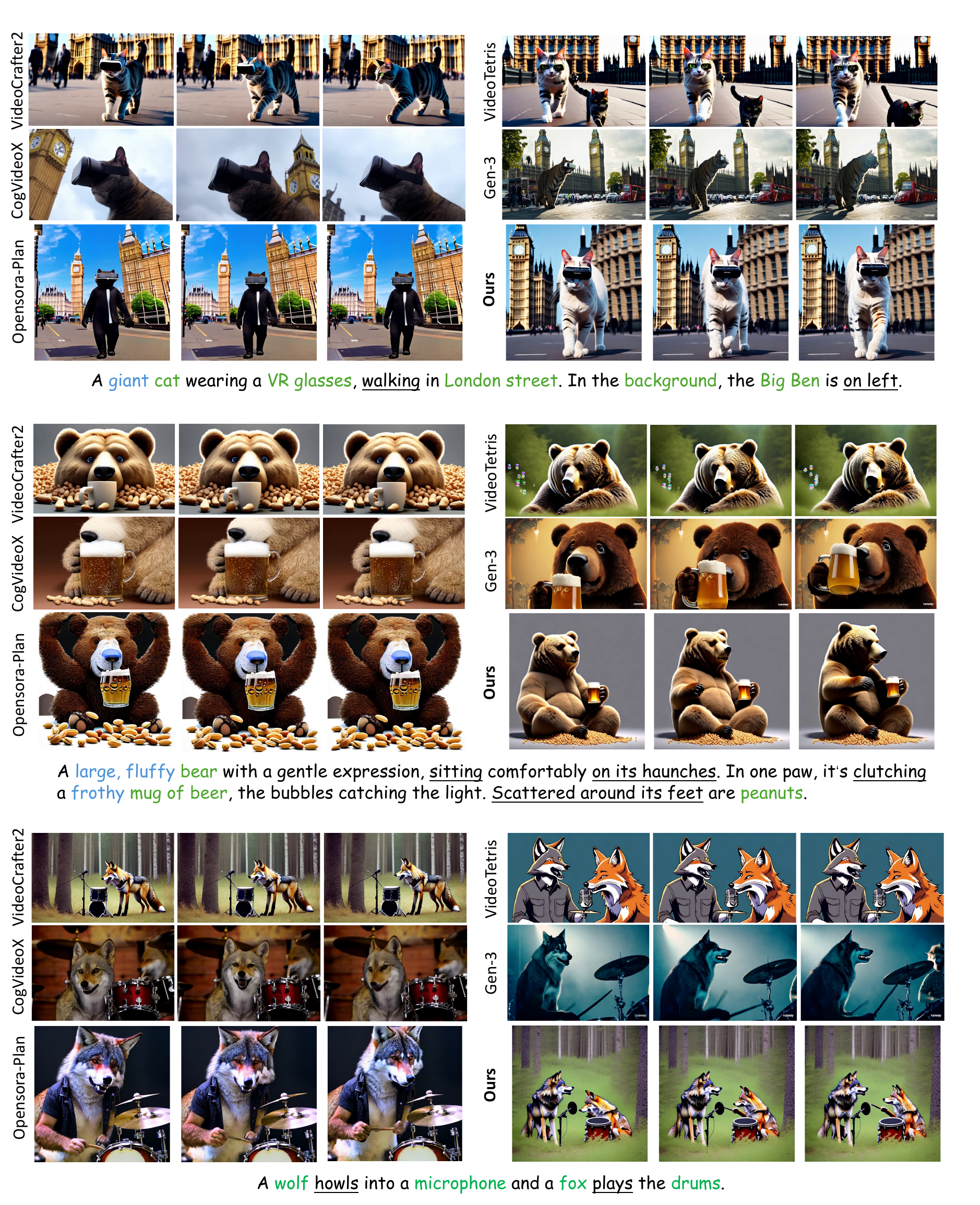}
   \caption{ 
   More qualitative comparisons. 
   }
   \label{fig:app_compare1}
\end{figure*}

\begin{figure*}[h]   %
  \centering
   \includegraphics[width=\linewidth]{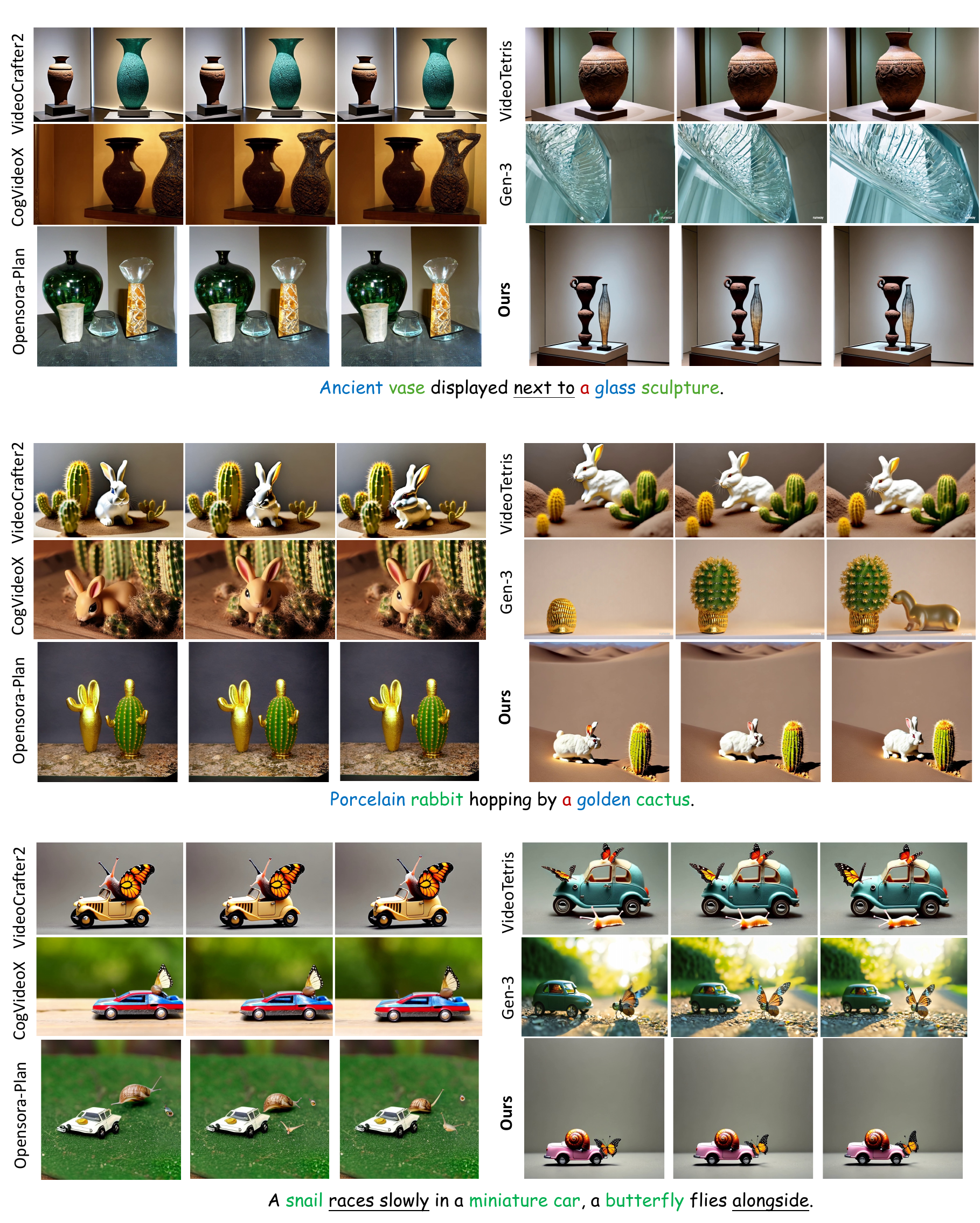}
   \caption{ 
   More qualitative comparisons. 
   }
   \label{fig:app_compare2}
\end{figure*}

We show visual comparisons on the video frames of our proposed method~\name~and VideoCrafter2~\cite{chen2024videocrafter2}, CogVideoX-5B~\cite{yang2024cogvideox}, Gen-3~\cite{gen3}, VideoTetris~\cite{videoteris}, and Open-Sora-Plan ~\cite{pku_yuan_lab_and_tuzhan_ai_etc_2024_10948109}. 

In the first example of~\Cref{fig:app_compare1} (the first row).
VideoCrafter2~\cite{chen2024videocrafter2} omits the VR glasses in the last frame.
CogVideoX~\cite{yang2024cogvideox} generates the video that the Big Ben is not always on the left side of the cat.
Open-Sora-Plan
~\cite{pku_yuan_lab_and_tuzhan_ai_etc_2024_10948109} 
depicts only a cat's head on a human body. VideoTetris~\cite{videoteris} generates two cats instead of one and omits the VR glasses, and Gen-3~\cite{gen3} does not follow the specified scene layout, generating two ``Big Ben'', one on the left and the other on the right.
In the second example of~\Cref{fig:app_compare1} (the second row), VideoTetris~\cite{videoteris} and Gen-3~\cite{gen3} do not depict ``sitting on haunches'', and Open-Sora-Plan ~\cite{pku_yuan_lab_and_tuzhan_ai_etc_2024_10948109} omits ``clutching a frothy mug of beer''. VideoCrafter2~\cite{chen2024videocrafter2} does not depict ``sitting on haunches'' with only the head of the bear. CogVideoX~\cite{yang2024cogvideox} only shows partial views of the bear.
In the third example of~\Cref{fig:app_compare1} (the third row), VideoCrafter2~\cite{chen2024videocrafter2}, Open-Sora-Plan ~\cite{pku_yuan_lab_and_tuzhan_ai_etc_2024_10948109}, and Gen-3~\cite{gen3} only generate ``one wolf'' instead of ``a wolf'' and ``a fox''. CogVideoX~\cite{yang2024cogvideox} fails to generate ``a microphone''
, while VideoTetris~\cite{videoteris} lacks the action ``plays the drum''.

As shown in~\Cref{fig:app_compare2}, all existing models fail to generate correct videos with the compositional prompts, including multiple objects, attribute binding, generative numeracy, and action binding. 
For example, in the first prompt in~\Cref{fig:app_compare2} (the first row), VideoCrafter2~\cite{chen2024videocrafter2}, CogVideoX~\cite{yang2024cogvideox}, VideoTetris~\cite{videoteris} and Open-Sora-Plan ~\cite{pku_yuan_lab_and_tuzhan_ai_etc_2024_10948109} fail to generate ``a glass sculpture'', while Gen-3~\cite{gen3} omits ``ancient vase''. 
For the second prompt in~\Cref{fig:app_compare2} (the second row), Gen-3~\cite{gen3} and Open-Sora-Plan ~\cite{pku_yuan_lab_and_tuzhan_ai_etc_2024_10948109} lack ``porcelain rabbit'' in some or all frames, while VideoCrafter2~\cite{chen2024videocrafter2}, CogVideoX~\cite{yang2024cogvideox}, and VideoTetris~\cite{videoteris} generate several ``golden cactus'' instead of ``one'' indicated in the prompt. 
For the third prompt in~\Cref{fig:app_compare2} (the third row), Open-Sora-Plan ~\cite{pku_yuan_lab_and_tuzhan_ai_etc_2024_10948109} omits the ``butterfly'', while Gen-3~\cite{gen3} fails to include the ``snail''. CogVideoX~\cite{yang2024cogvideox} and VideoTetris~\cite{videoteris} do not accurately depict the action ``a snail races in a miniature car'', and in VideoCrafter2~\cite{chen2024videocrafter2}, the wings of the butterfly appear to grow on the snail.

These examples demonstrate the challenges of compositional text-to-video generation faced by both the open-source and the commercial models. 
Our proposed ~\name{} correctly reflect the composition of multiple objects, attribute binding, generative numeracy, showing advantages in compositionality.

\subsection{Qualitative Results}
\label{app: qualitative_ours}
\begin{figure*}[h]   %
  \centering
   \includegraphics[width=\linewidth]{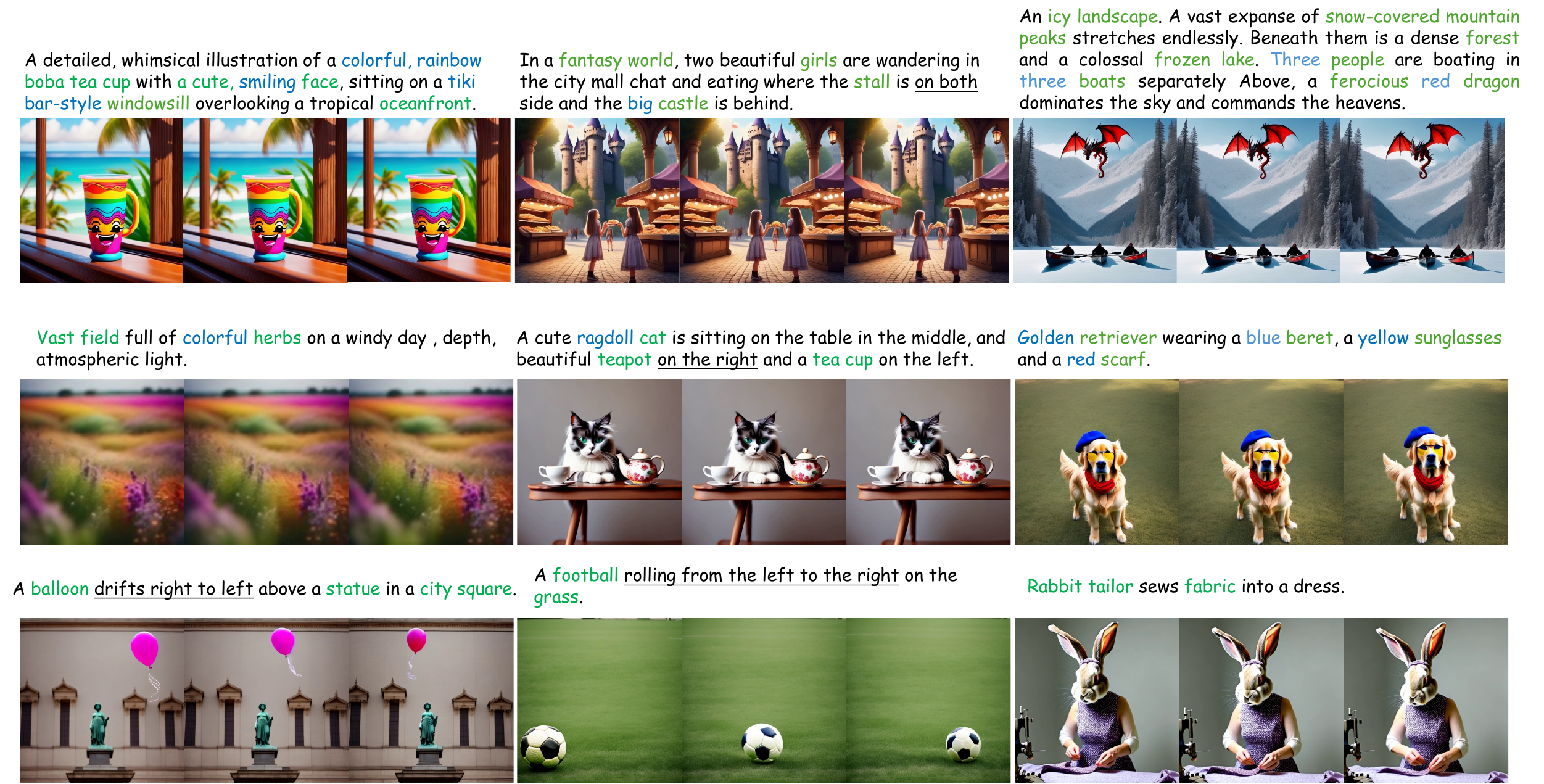}
   \caption{ 
   \textbf{Qualitative results} of ~\name{}. ~\name{} shows ability to adhere to complex compositional prompts, including attribute binding for multiple objects, temporal dynamics for object movement, and interactions.}
   \label{fig:app_qualitative_1}
\end{figure*}

\begin{figure*}[h]   %
  \centering
   \includegraphics[width=\linewidth]{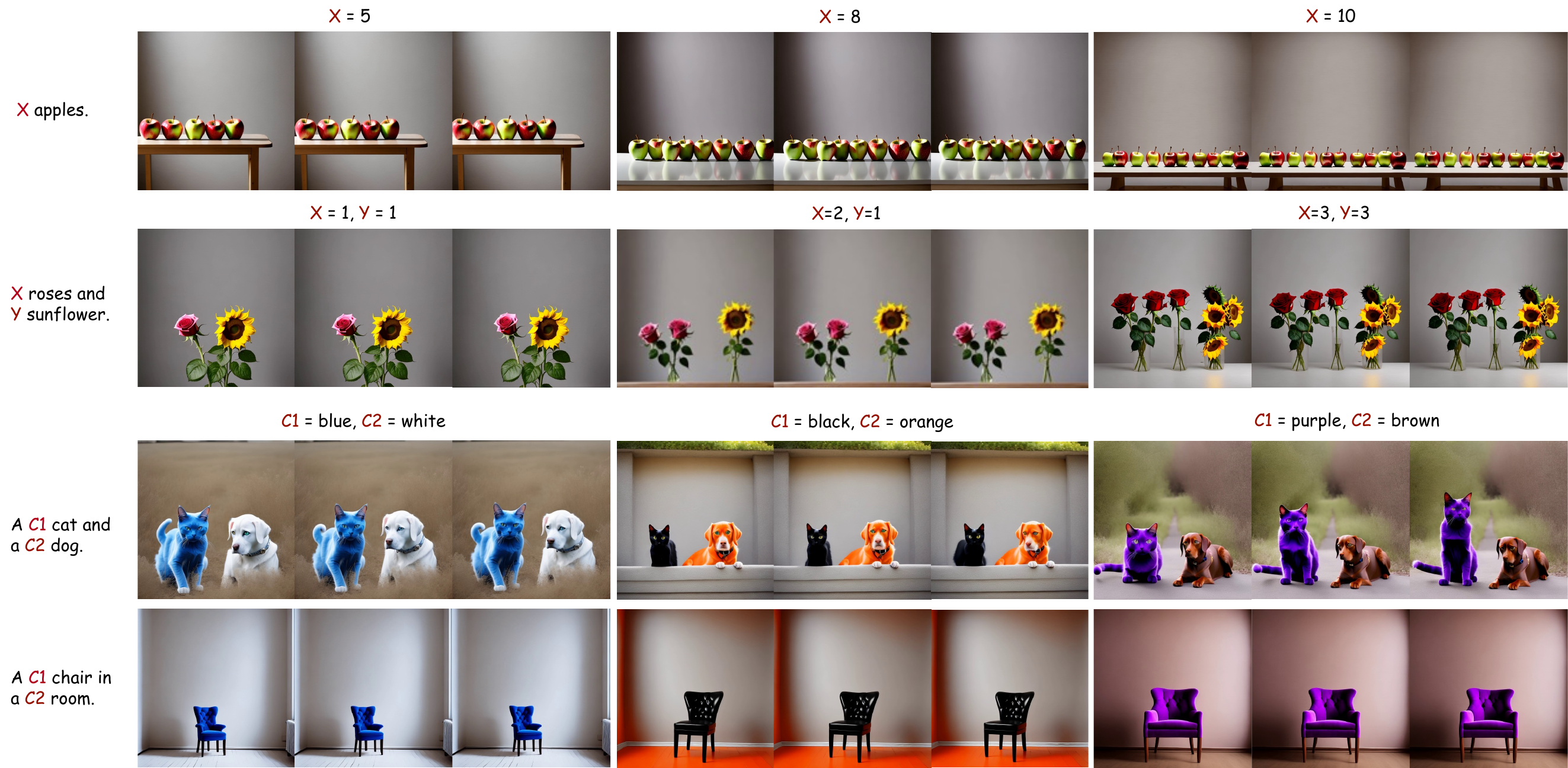}
   \caption{ 
\textbf{Qualitative results} of ~\name{}. ~\name{} shows exhibit superior performances in controllability of generative numeracy, multiple objects with different attributes in compositionality.}

   \label{fig:app_qualitative_2}
\end{figure*}

We show qualitative results in~\Cref{fig:app_qualitative_1} and ~\Cref{fig:app_qualitative_2}. 
In ~\Cref{fig:app_qualitative_1}, our proposed ~\name{} show ability to adhere to complex compositional prompts, including attribute binding for multiple objects, temporal dynamics for object movement, and interactions.
In ~\Cref{fig:app_qualitative_2}, we show the qualitative results in various settings of generative numeracy, multiple objects with different attribute bindings, indicating that our proposed ~\name{} exhibit superior performances in controllability of compositionality.

\subsection{Results on Iterative Generation}
\label{app: iteration_results}

We provide visual examples in~\Cref{fig:app_qualitative} to illustrate how our multi-agent framework works. In the case of~\Cref{fig:qualitative_1}, the \designagent{} creates an initial layout for the rope and boat, but the first generated video lacks the implied "tug" motion. In the \redesign{} stage, \alignmentagent{} detects this absence of interaction, and \suggagent{} proposes adjusting the bounding boxes to create tension between the objects. The \refineagent{} then adjusts the bounding boxes, and the \structuredagent{} standardizes the guidance, resulting in a refined video that aligns with the prompt.

In~\Cref{fig:qualitative_3}, despite the corrected guidance from the \design{} stage, the initial generated video still exhibits mismatches in both object quantity and motion direction. Over the next two iterations, agents in the \redesign{} stage progressively increase the guidance scale of the car while jointly refining both the motion direction and object quantity to achieve alignment with the prompt.

\subsection{Analysis of Various Guidance Settings}
We provide an analysis of various guidance settings, including structured layout, guidance scale, and new text prompt in T2V-CompBench~\cite{sun2024t2vcompbench} in~\Cref{fig:multi_guidance}. 
The number of corrections across iterations from one to five is presented in~\Cref{fig:multi_guidance_num}. Since only the structured layout is provided during the \design{} stage, with the guidance scale and text prompt set to default values, we attribute the corrections entirely to the structured layout (iteration one).
With the \redesign{} stage engaged (from iteration two to five), the overall trend of all guidance shows a decline. The guidance scale contributes most to the corrections across iterations two to four, while the contributions of structured layout and new text prompt vary with iterations.

The contribution (\%) of different guidance types to the video scores with the \design{} and \redesign{} stages is depicted in~\Cref{fig:multi_guidance_w_iter1}, while with only the \redesign{} stage is shown in~\Cref{fig:multi_guidance_wo_iter1}. With both \design{} and \redesign{} stages, structured layout contributes up to 80.4\%, followed by new text prompt (12.6\%) and guidance scale (7.0\%). With only the \redesign{} stage, the contributions of various guidance are relatively balanced, with the new text prompt contributing slightly more than the structured layout, followed by the guidance scale.

\section{Limitation and Potential Negative Social Impacts}
\label{app: discusstion}
Our method \name{} employs MLLMs as multi-agent for compositional video generation. Although \name{} shows substantial enhancement over existing methods in compositional text-to-video generation, there is still potential for further improvement.
The method depends on the performance of the MLLMs used. Here we adopt GPT-4o as MLLM, for those tasks that exceed the capability of GPT-4o, our method may fail. 
Besides, \name{} inherits limitations from the base generation model when it comes to generating objects or actions it struggles with.

For the potential negative social impacts, the community must recognize the impacts that can result from the misuse of video generation models. These impacts include the creation of misleading or harmful content, which could intensify challenges such as the spread of misinformation and the proliferation of deepfakes.

\begin{table*}[h]
\caption{Our full example for~\Cref{fig:app_qualitative_2}, with the prompt ``A car driving right to left on the moon''.}
\label{tab:app_full_example3}
\resizebox{\linewidth}{!}{
\begin{tabular}{@{}ll@{}}

\toprule
\textbf{\design{} stage} &  \\ \hline
Design & \begin{tabular}[c]{@{}l@{}} Reasoning: The car is driving from right to left, so its x-coordinate should decrease while its y-coordinate remains constant. \\The moon's surface is flat and has low gravity, so the car's movement will be smooth and consistent. \\
Frame 1: \texttt{{[}\{'id': 0, 'name': 'car', 'box': {[}400, 350, 100, 50{]}\}{]}} \\
Frame 2: \texttt{{[}\{'id': 0, 'name': 'car', 'box': {[}320, 350, 100, 50{]}\}{]}} \\
Frame 3: \texttt{{[}\{'id': 0, 'name': 'car', 'box': {[}240, 350, 100, 50{]}\}{]}} \\
Frame 4: \texttt{{[}\{'id': 0, 'name': 'car', 'box': {[}160, 350, 100, 50{]}\}{]}} \\
Frame 5: \texttt{{[}\{'id': 0, 'name': 'car', 'box': {[}80, 350, 100, 50{]}\}{]}} \\
Frame 6: \texttt{{[}\{'id': 0, 'name': 'car', 'box': {[}0, 350, 100, 50{]}\}{]}} \\
Background keyword: moon \\
Generation suggestion: None \\
New prompt: A car drives from right to left on the moon's surface. \\
\end{tabular}\\

\midrule

\textbf{\redesign{} stage} & \textbf{iter 1} \\ \hline
  & \textbf{\alignmentagent{}} \\ 

Verification & \begin{tabular}[c]{@{}l@{}} 
The alignment check reveals the following issues: \\
1. Quantity of Objects: There are two cars present in the video instead of one. \\
2. Motion Direction: The cars are not clearly shown moving from right to left. \\
3. Correctness of Attributes: The cars appear to be on the moon, which aligns with the prompt. \\
Overall, there is a mismatch in the quantity of objects and the motion direction. 
\end{tabular}\\

  & \textbf{\suggagent{}} \\ 
Suggestion & \begin{tabular}[c]{@{}l@{}} 
1. Suggest corrections for the bounding boxes: \\
- Remove one of the cars to correct the quantity issue. \\
- Ensure the remaining car is clearly shown moving from right to left by adjusting its position across frames to depict motion. \\ 
2. Choose the suitable correction agent: B1. (\refineagent{} for spatial dynamics)\\
\end{tabular}  \\

 & \textbf{\refineagent{}} \\ 

Correction & \begin{tabular}[c]{@{}l@{}} 
The correction suggestion indicates that one of the cars should be removed to correct the quantity issue. \\The remaining car should be shown moving from right to left, which aligns with the previous bounding boxes. \\
Here are the corrected bounding boxes: \\
Corrected Bounding Boxes: \\
- Frame 1: [{'id': 0, 'name': 'car', 'box': [400, 350, 100, 50]}] 
- Frame 2: [{'id': 0, 'name': 'car', 'box': [320, 350, 100, 50]}] \\
- Frame 3: [{'id': 0, 'name': 'car', 'box': [240, 350, 100, 50]}] 
- Frame 4: [{'id': 0, 'name': 'car', 'box': [160, 350, 100, 50]}] \\
- Frame 5: [{'id': 0, 'name': 'car', 'box': [80, 350, 100, 50]}] 
- Frame 6: [{'id': 0, 'name': 'car', 'box': [0, 350, 100, 50]}] \\
Explanation: \\
The bounding boxes remain the same as the previous ones, as they already depict the car moving from right to left. \\The suggestion to generation is to emphasize the movement of the single car across the frames. 
\end{tabular}\\

 & \textbf{\structuredagent{}} \\ 
Output & \begin{tabular}[c]{@{}l@{}} Reasoning: The sequence depicts a single car moving from right to left across the frames. \\Initially, the car is positioned on the right side of the frame and gradually moves to the left, \\
maintaining a consistent y-coordinate, which aligns with the prompt of a car driving on the moon. \\
Frame 1: \texttt{{[}\{'id': 0, 'name': 'car', 'box': {[}400, 350, 100, 50{]}\}{]}} \\ 
Frame 2: \texttt{{[}\{'id': 0, 'name': 'car', 'box': {[}320, 350, 100, 50{]}\}{]}} \\ 
Frame 3: \texttt{{[}\{'id': 0, 'name': 'car', 'box': {[}240, 350, 100, 50{]}\}{]}} \\ 
Frame 4: \texttt{{[}\{'id': 0, 'name': 'car', 'box': {[}160, 350, 100, 50{]}\}{]}} \\ 
Frame 5: \texttt{{[}\{'id': 0, 'name': 'car', 'box': {[}80, 350, 100, 50{]}\}{]}} \\ 
Frame 6: \texttt{{[}\{'id': 0, 'name': 'car', 'box': {[}0, 350, 100, 50{]}\}{]}} \\ 
Background keyword: moon \\ 
Generation suggestion: emphasize id 0 \\ 
New prompt: A car driving right to left on the moon. 
 \end{tabular}\\ 

\hline
 & \textbf{iter 2} \\ \hline
& \textbf{\alignmentagent{}} \\ 
Verification & \begin{tabular}[c]{@{}l@{}} 
The alignment check reveals the following issues:\\
Direction of Motion: The car is moving from left to right, which is opposite to the prompt's description of right to left.\\
Existence of Specified Objects: The car and the moon surface are present, which aligns with the prompt.\\
Quantity of Objects: The single car is correctly depicted.\\
Correctness of Object Attributes: The car and the lunar surface appear as expected.\\
Overall, the main issue is the direction of the car's movement.\\
Alignment Check Result: The video does not fully align with the prompt due to the incorrect motion direction of the car.
\end{tabular}\\

  & \textbf{\suggagent{}} \\ 
Suggestion & \begin{tabular}[c]{@{}l@{}} 
1. Suggest corrections for the bounding boxes: Adjust the motion path of the car to reflect a right-to-left direction instead of left-to-right.\\
2. Choose the suitable correction agent: B1. (\refineagent{} for spatial dynamics) 
\end{tabular}\\

  & \textbf{\refineagent{}} \\ 
  Correction & \begin{tabular}[c]{@{}l@{}}
Based on the correction suggestion, the bounding boxes need to be adjusted to reflect the car moving from right to left. Here are the corrected bounding boxes:\\
**Corrected Bounding Boxes:**\\
- **Frame 1:** [{'id': 0, 'name': 'car', 'box': [400, 350, 100, 50]}] - **Frame 2:** [{'id': 0, 'name': 'car', 'box': [320, 350, 100, 50]}] \\- **Frame 3:** [{'id': 0, 'name': 'car', 'box': [240, 350, 100, 50]}]- **Frame 4:** [{'id': 0, 'name': 'car', 'box': [160, 350, 100, 50]}]\\- **Frame 5:** [{'id': 0, 'name': 'car', 'box': [80, 350, 100, 50]}]- **Frame 6:** [{'id': 0, 'name': 'car', 'box': [0, 350, 100, 50]}]\\
**Comparison and Suggestion:**
The corrected bounding boxes are the same as the previous ones, indicating that the car is already moving from right to left as intended. \\
**Suggestion to Generation:**
Emphasize the car (id 0) to ensure it stands out against the moon background.
\end{tabular}\\ 

  & \textbf{\structuredagent{}} \\ 

Output & \begin{tabular}[c]{@{}l@{}}  Reasoning:  The sequence depicts a car moving from right to left across the frames. The car's x-coordinate decreases consistently, indicating its movement from right to left \\as described in the prompt. \\
The background is the moon, providing a unique setting for the car's journey. \\
Frame 1: \texttt{{[}\{'id': 0, 'name': 'car', 'box': {[}400, 350, 100, 50{]}\}{]}} \\ 
Frame 2: \texttt{{[}\{'id': 0, 'name': 'car', 'box': {[}320, 350, 100, 50{]}\}{]}} \\ 
Frame 3: \texttt{{[}\{'id': 0, 'name': 'car', 'box': {[}240, 350, 100, 50{]}\}{]}} \\ 
Frame 4: \texttt{{[}\{'id': 0, 'name': 'car', 'box': {[}160, 350, 100, 50{]}\}{]}} \\
Frame 5: \texttt{{[}\{'id': 0, 'name': 'car', 'box': {[}80, 350, 100, 50{]}\}{]}} \\ 
Frame 6: \texttt{{[}\{'id': 0, 'name': 'car', 'box': {[}0, 350, 100, 50{]}\}{]}} \\ 
Background keyword: moon \\ 
Generation Suggestion : emphasize id 0. \\ 
New Prompt: A car driving right to left on the moon. \\ 
\end{tabular}\\

\bottomrule
\end{tabular}
}
\end{table*}

\begin{table*}[h]
\caption{Our full example for~\Cref{fig:qualitative_2} in the main paper, with the prompt ``Rabbit police officer directs traffic''.}
\label{tab:app_full_example2}
\resizebox{\linewidth}{!}{
\begin{tabular}{@{}ll@{}}
\toprule

\textbf{\design{} stage} &  \\ \hline
Design & \begin{tabular}[c]{@{}l@{}} Reasoning: The rabbit police officer will likely be standing in one place, directing traffic, so its position will remain relatively stable across frames. \\
Frame 1: \texttt{{[}\{'id': 0, 'name': 'rabbit police officer', 'box': {[}206, 256, 100, 150{]}\}{]}} \\
Frame 2: \texttt{{[}\{'id': 0, 'name': 'rabbit police officer', 'box': {[}206, 256, 100, 150{]}\}{]}} \\
Frame 3: \texttt{{[}\{'id': 0, 'name': 'rabbit police officer', 'box': {[}206, 256, 100, 150{]}\}{]}} \\
Frame 4: \texttt{{[}\{'id': 0, 'name': 'rabbit police officer', 'box': {[}206, 256, 100, 150{]}\}{]}} \\ 
Frame 5: \texttt{{[}\{'id': 0, 'name': 'rabbit police officer', 'box': {[}206, 256, 100, 150{]}\}{]}} \\
Frame 6: \texttt{{[}\{'id': 0, 'name': 'rabbit police officer', 'box': {[}206, 256, 100, 150{]}\}{]}} \\ 
Background keyword: street \\ 
New prompt: A rabbit police officer directing traffic on the street.
 \end{tabular}\\

\midrule

\textbf{\redesign{} stage} & \textbf{iter 1} \\ \hline
  & \textbf{\alignmentagent{}} \\ 

Verification & \begin{tabular}[c]{@{}l@{}l@{}l@{}l@{}l@{}} The alignment check reveals the following issues: \\1. **Existence of Specified Objects**: The rabbit is present and dressed as a police officer. \\ 2. **Quantity of Objects**: There is one rabbit police officer. \\ 3. **Correctness of Object Attributes**: The rabbit is wearing a police uniform, which aligns with the prompt.\\ 4. **Accuracy of Relationships**: The rabbit is not shown directing traffic in the frames provided. \\ Overall, the video does not fully align with the prompt as the action of directing traffic is missing. \end{tabular} \\ 

  & \textbf{\suggagent{}} \\ 
Suggestion & \begin{tabular}[c]{@{}l@{}} 1. **Suggest corrections for the bounding boxes**: \\ - Add a bounding box to include a scene or action where the rabbit is directing traffic. This could involve positioning the rabbit with an arm raised or using a gesture to indicate traffic direction. \\ 2. **Choose the suitable correction agent**: A. (\refineagent{} for consistency)\\ The focus is on maintaining the consistent attributes of the rabbit as a police officer while introducing the fixed spatial relationship of directing traffic. \end{tabular} \\

 & \textbf{\refineagent{}} \\ 
 
Correction & \begin{tabular}[c]{@{}l@{}} **Corrected Bounding Boxes:** \\ 1. **Frame 1:** \\ - [{'id': 0, 'name': 'rabbit police officer', 'box': [206, 256, 100, 150]}] \\ 2. **Frame 2:** \\ - [{'id': 0, 'name': 'rabbit police officer', 'box': [206, 256, 100, 150]}] \\ 3. **Frame 3:** \\ - [{'id': 0, 'name': 'rabbit police officer', 'box': [206, 256, 100, 150]}] \\ 4. **Frame 4:** \\ - [{'id': 0, 'name': 'rabbit police officer', 'box': [206, 256, 100, 150]}] \\ 5. **Frame 5:** \\ - [{'id': 0, 'name': 'rabbit police officer', 'box': [206, 256, 100, 150]}] \\ 6. **Frame 6:** \\ - [{'id': 0, 'name': 'rabbit police officer', 'box': [206, 256, 100, 150]}] \\ **Explanation:** \\ - The bounding boxes remain the same as the previous ones since the suggestion primarily involves adding an action or gesture, which is not reflected in the bounding box dimensions. \\ - The rabbit police officer's position is stable, and the box size is appropriate for visibility. \\ **Suggestion to Generation:** \\ - Emphasize the rabbit police officer's gesture or action of directing traffic within the existing bounding box. \end{tabular} \\

 & \textbf{\structuredagent{}} \\ 
Output & \begin{tabular}[c]{@{}l@{}} Reasoning: The video features a rabbit police officer consistently positioned in the same location across all frames, with no additional objects or changes in state. \\The prompt needs to be rephrased to include all objects, but since only one object is present, the original prompt is sufficient. \\
Frame 1: \texttt{{[}\{'id': 0, 'name': 'rabbit police officer', 'box': {[}206, 256, 100, 150{]}\}{]}} \\
Frame 2: \texttt{{[}\{'id': 0, 'name': 'rabbit police officer', 'box': {[}206, 256, 100, 150{]}\}{]}} \\
Frame 3: \texttt{{[}\{'id': 0, 'name': 'rabbit police officer', 'box': {[}206, 256, 100, 150{]}\}{]}}\\
Frame 4: \texttt{{[}\{'id': 0, 'name': 'rabbit police officer', 'box': {[}206, 256, 100, 150{]}\}{]}} \\ 
Frame 5: \texttt{{[}\{'id': 0, 'name': 'rabbit police officer', 'box': {[}206, 256, 100, 150{]}\}{]}} \\ 
Frame 6: \texttt{{[}\{'id': 0, 'name': 'rabbit police officer', 'box': {[}206, 256, 100, 150{]}\}{]}} \\ 
Background keyword: street \\ 
Generation suggestion: emphasize id 0 \\ 
New prompt: Rabbit police officer directs traffic. \\ 
\end{tabular}\\

\hline
 & \textbf{iter 2} \\ \hline
& \textbf{\alignmentagent{}} \\

Verification & \begin{tabular}[c]{@{}l@{}}
The video does not align well with the prompt. \\
Detailed alignment check results: \\
- The specified object, "rabbit police officer," is not clearly depicted as a police officer. \\
There are no visible indicators (e.g., uniform, badge) that suggest the rabbit is a police officer. \\
- The action "directs traffic" is not shown. The rabbit is simply standing on a path without any indication of directing traffic. \\
Overall, the video lacks the necessary attributes and actions described in the prompt. 
\end{tabular}\\

  & \textbf{\suggagent{}} \\ 
Corrections & \begin{tabular}[c]{@{}l@{}} (1) Suggest corrections for the bounding boxes: \\
 - Add a bounding box around the rabbit and include elements that indicate it is a police officer, such as a small uniform or badge. \\
 - Introduce additional elements or objects within the scene to depict traffic (e.g., small toy cars) and position them in a way that suggests the rabbit is directing them. \\
 - Ensure the rabbit's posture or gestures imply directing traffic, such as an arm raised or a whistle. \\
 (2) Choose the suitable correction agent: A. (\refineagent{} for consistency)
 \end{tabular}\\ 

  & \textbf{\refineagent{}} \\ 
Correction & \begin{tabular}[c]{@{}l@{}l@{}}
Corrected Bounding Boxes: \\
Frame 1: Rabbit police officer: [200, 250, 112, 162], Toy car 1: [50, 400, 60, 30], Toy car 2: [400, 400, 60, 30] \\
Frame 2: 
Rabbit police officer: [200, 250, 112, 162], 
Toy car 1: [50, 400, 60, 30],
Toy car 2: [400, 400, 60, 30] \\
Frame 3: 
Rabbit police officer: [200, 250, 112, 162],
Toy car 1: [50, 400, 60, 30],
Toy car 2: [400, 400, 60, 30] \\
Frame 4: 
Rabbit police officer: [200, 250, 112, 162],
Toy car 1: [50, 400, 60, 30],
Toy car 2: [400, 400, 60, 30] \\
Frame 5: 
Rabbit police officer: [200, 250, 112, 162],
Toy car 1: [50, 400, 60, 30],
Toy car 2: [400, 400, 60, 30] \\
Frame 6:
Rabbit police officer: [200, 250, 112, 162],
Toy car 1: [50, 400, 60, 30],
Toy car 2: [400, 400, 60, 30]\\
Differences and Suggestions: 
Differences: Additional bounding boxes for toy cars have been added to depict traffic, and the rabbit's bounding box has been slightly adjusted for better coverage. \\
Suggestion to Generation: Emphasize the rabbit police officer's directing posture and ensure the toy cars are clearly visible in the scene. \\
\end{tabular}\\  

  & \textbf{\structuredagent{}} \\

Output & \begin{tabular}[c]{@{}l@{}} 
Reasoning: The scene involves a rabbit police officer consistently directing traffic, with two toy cars present throughout the sequence. The rabbit's position remains constant, \\while the toy cars are positioned at opposite ends of the frame, maintaining their positions to depict a traffic scenario. \\
Frame 1: \texttt{{[}\{'id': 0, 'name': 'rabbit police officer', 'box': {[}200, 250, 112, 162{]}\}, \{'id': 1, 'name': 'toy car 1', 'box': {[}50, 400, 60, 30{]}\},}\\
\texttt{{[}\{'id': 2, 'name': 'toy car 2', 'box': {[}400, 400, 60, 30{]}\}{]}} \\
Frame 2: \texttt{{[}\{'id': 0, 'name': 'rabbit police officer', 'box': {[}200, 250, 112, 162{]}\}, \{'id': 1, 'name': 'toy car 1', 'box': {[}50, 400, 60, 30{]}\}, }\\
\texttt{{[}\{'id': 2, 'name': 'toy car 2', 'box': {[}400, 400, 60, 30{]}\}{]}} \\
Frame 3: \texttt{{[}\{'id': 0, 'name': 'rabbit police officer', 'box': {[}200, 250, 112, 162{]}\}, \{'id': 1, 'name': 'toy car 1', 'box': {[}50, 400, 60, 30{]}\},}\\
\texttt{{[}\{'id': 2, 'name': 'toy car 2', 'box': {[}400, 400, 60, 30{]}\}{]}} \\
Frame 4: \texttt{{[}\{'id': 0, 'name': 'rabbit police officer', 'box': {[}200, 250, 112, 162{]}\}, \{'id': 1, 'name': 'toy car 1', 'box': {[}50, 400, 60, 30{]}\},}\\
\texttt{{[}\{'id': 2, 'name': 'toy car 2', 'box': {[}400, 400, 60, 30{]}\}{]}} \\
Frame 5: \texttt{{[}\{'id': 0, 'name': 'rabbit police officer', 'box': {[}200, 250, 112, 162{]}\}, \{'id': 1, 'name': 'toy car 1', 'box': {[}50, 400, 60, 30{]}\},}\\
\texttt{{[}\{'id': 2, 'name': 'toy car 2', 'box': {[}400, 400, 60, 30{]}\}{]}} \\
Frame 6: \texttt{{[}\{'id': 0, 'name': 'rabbit police officer', 'box': {[}200, 250, 112, 162{]}\}, \{'id': 1, 'name': 'toy car 1', 'box': {[}50, 400, 60, 30{]}\}, }\\
\texttt{{[}\{'id': 2, 'name': 'toy car 2', 'box': {[}400, 400, 60, 30{]}\}{]}} \\
Background keyword: street \\
Generation suggestion: emphasize id 0 \\
New prompt: A rabbit police officer directs traffic with two toy cars on the street. \\
\end{tabular}\\

\bottomrule
\end{tabular}
}
\end{table*}

\end{document}